\pdfoutput=1

\documentclass[11pt]{article}

\usepackage[]{acl}

\usepackage{times}
\usepackage{latexsym}
\usepackage{xcolor}
\usepackage{pgfplots}

\usepackage{times}
\usepackage{svg}
\usepackage{latexsym}
\usepackage{enumitem}

\usepackage{booktabs}
\usepackage{multirow}
  
\usepackage{microtype}
\usepackage{csquotes}
\usepackage{bm}
\usepackage{amsmath,amssymb, amsfonts,mathtools, bm}
\usepackage{soul}
\usepackage{adjustbox}
\usepackage{resizegather}
\usepackage{appendix}
\usepackage{etoolbox}
\usepackage{caption}
\usepackage{subcaption}
\usepackage{enumerate}
\usepackage{graphicx}
\usepackage{float}
\usepackage{scalerel}
\usepackage{svg}
\BeforeBeginEnvironment{appendices}{\clearpage}
\usepackage[ruled,vlined]{algorithm2e}
\usepackage{linguex}
\usepackage[utf8]{inputenc}
\usepackage[font=small,labelfont=bf]{caption}
\usepackage{cleveref}
\Crefname{ALC@unique}{Line}{Lines}
\Crefname{section}{\S}{\S\S}
\Crefname{section}{\S}{\S\S}
\Crefformat{section}{\S#2#1#3}
\Crefname{table}{Table}{Tables}
\Crefname{figure}{Fig.}{Fig.}
\Crefname{algorithm}{Alg}{Alg}
\Crefname{algorithm}{Alg}{Alg}
\Crefname{line}{line}{lines}
\Crefname{appendix}{\S\!\!}{\S\!\!}
\Crefname{thm}{Theorem}{}
\Crefname{prop}{Prop.\@}{Props.\@}
\Crefname{defin}{Definition}{Definitions}
\Crefname{lemma}{Lemma}{Lemmata}
\Crefname{cor}{Corollary}{Corollaries}
\Crefname{equation}{}{}
\Crefname{myexample}{Example}{Examples}

\DeclareMathOperator*{\argmax}{argmax}

\usepackage{relsize}
\usepackage{booktabs}
\usepackage{makecell}
\usepackage{relsize}

\usepackage{array}
\usepackage{ragged2e}
\newcolumntype{P}[1]{>{\RaggedRight\hspace{0pt}}p{#1}}
\newcolumntype{X}[1]{>{\RaggedRight\hspace*{0pt}}p{#1}}
\usepackage{tikz}
\usetikzlibrary{backgrounds}
\usetikzlibrary{arrows,shapes}
\usetikzlibrary{tikzmark}
\usetikzlibrary{calc}
\usetikzlibrary{arrows,shapes,positioning,shadows,trees,mindmap}
\usepackage[edges]{forest}
\usetikzlibrary{arrows.meta}
\colorlet{linecol}{black!75}
\usepackage{xkcdcolors} %
\usepackage{tcolorbox}

\definecolor{darkorange}{rgb}{1, 0.549, 0}

\usepackage{stmaryrd}

\usepackage{stmaryrd}

\usepackage{mathtools}
\usepackage{todonotes}
\makeatletter
\newcommand*\iftodonotes{\if@todonotes@disabled\expandafter\@secondoftwo\else\expandafter\@firstoftwo\fi}  %
\makeatother

\usepackage{enumitem}

\usepackage[T1]{fontenc}

\usepackage[utf8]{inputenc}

\usepackage{microtype}

\usepackage[english]{babel}
\usepackage{amsmath,amssymb}
\usepackage{tabularx}
\usepackage{booktabs}
\usepackage{enumitem}
\usepackage{graphicx}
\usepackage{multirow}

\title{
Opportunities and Challenges in Neural Dialog Tutoring
}

\author{
    Jakub Macina$^{\ast 1, 2}$ \quad
    Nico Daheim$^{\ast 3}$ \quad
    Lingzhi Wang$^4$ \\
    \textbf{
    Tanmay Sinha$^{5}$ \quad
    Manu Kapur$^5$ \quad
    Iryna Gurevych$^3$ \quad
    Mrinmaya Sachan$^{1}$
    } \\ \text{} \\
  $^{1}$Department of Computer Science, ETH Zürich \quad
  $^2$ ETH AI Center \\
  $^{3}$ Ubiquitous Knowledge Processing Lab (UKP Lab), Department of Computer Science\\ and Hessian Center for AI (hessian.AI), TU Darmstadt \\
  $^4$ The Chinese University of Hong Kong \\
  $^5$Professorship for Learning Sciences and Higher Education, ETH Zürich \\
  \texttt{jakub.macina@ai.ethz.ch}
}

\begin{document}
\maketitle
\def\thefootnote{*}\footnotetext{
Equal contribution.}\def\thefootnote{\arabic{footnote}}
\begin{abstract}
Designing dialog tutors has been challenging as it involves modeling the diverse and complex pedagogical strategies employed by human tutors. Although there have been significant recent advances in neural conversational systems using large language models (LLMs) and growth in available dialog corpora, dialog tutoring has largely remained unaffected by these advances.
In this paper, we rigorously analyze various generative language models on two dialog tutoring datasets for language learning using automatic and human evaluations to understand the new opportunities brought by these advances as well as the challenges we must overcome to build models that would be usable in real educational settings. %
We find that although current approaches can model tutoring in constrained learning scenarios when the number of concepts to be taught and possible teacher strategies are small, they perform poorly in less constrained scenarios.
Our human quality evaluation shows that both models and ground-truth annotations exhibit low performance in terms of equitable tutoring, which measures learning opportunities for students and how engaging the dialog is.
To understand the behavior of our models in a real tutoring setting, we conduct a user study using expert annotators and find a significantly large number of model reasoning errors in 45\% of conversations.
Finally, we connect our findings to outline future work.  
\end{abstract}

\hspace{.5em}\includegraphics[width=1.25em,height=1.25em]{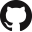}\hspace{.75em}\parbox{\dimexpr\linewidth-2\fboxsep-2\fboxrule}{\url{https://github.com/eth-nlped/dialog-tutoring}}
\vspace{-.5em}

\section{Introduction}

The goal of dialog tutoring research is to build systems that can tutor students using natural language conversation %
\cite{wollny2021we}. 
For several decades, learning scientists have been studying the features of domain-specific dialog tutoring systems that engender learning in students \citep{chi1994eliciting,graesser1995collaborative,moore2004generating,litman2006spoken,graesser2016conversations,ruan2019quizbot} and have established strong learning gains that are even comparable to human tutoring in specific domains \citep{nye2014autotutor}.
However, these systems require extensive authoring of materials by teachers \citep{maclellan2020domain} and therefore cannot fully utilize the scalability of online learning.

Building dialog tutors is technically challenging as tutoring dialogs typically exhibit properties that are absent in other forms of dialog. Tutoring dialogs are often \textit{long}, enabling students to be exposed to the concepts in a way that they can use them in future \citep{chi2014icap}, and \textit{grounded} in the learning scenarios \citep{graesser2009meta}. Finally, good dialog tutors are engaging and create opportunities to learn, providing students space to seek and provide explanations, and self-reflect \cite{chi2014icap,reiser_scaffolding_2004}.

The growing success of deep neural network based language generators in other dialog settings \citep{adiwardana2020towards,roller2020recipes} suggests new possibilities in dialog tutoring that could scale beyond domain-specific approaches. However, despite their promise, advances in neural generative models have seen little adoption in dialog tutoring.

In this paper, we 
contribute a
comprehensive study of the applicability of neural generative models in tutoring.
We formally introduce the dialog tutoring task and analyze existing tutoring datasets (\cref{sec:dialogtask}). Then,
we describe several generative and retrieval-based models for dialog tutoring (\cref{methods}) and benchmark them on two open-access dialog tutoring datasets for language learning: \textit{CIMA} \citep[][a crowdsourced role-played dataset for learning prepositional phrases in Italian]{stasaski-etal-2020-cima} and \textit{Teacher-Student Chatroom Corpus (TSCC)} \citep[][a one-to-one English tutoring dataset from an online chatroom]{caines2020teacher} (\cref{comparison-section}).
We evaluate our models on various automatic metrics (\cref{sec:eval_metrics}) as well as two human evaluation studies: an evaluation of the quality of the generated response with respect to various measures of goodness (\cref{human-quality}), as well as a more realistic user study with a learning interface (\cref{user-study-section}).

Overall, while we find that pretrained models improve over simpler baselines in terms of automatic metrics, our consequent human evaluation reveals several shortcomings that ought to be addressed before these models can be adopted in the real world.
We find that while neural generative models can model more constrained learning settings well, they struggle when the learning goal is more open-ended. %
Specifically, these models are unable to understand and
reason about student solutions and misconceptions, and thus, are unable to use effective pedagogical strategies.

We find that the field of dialog tutoring is significantly limited by the quantity and quality of available datasets. The available datasets are both too small and not rich enough to capture the nuances of the dialog tutoring problem.
Our analysis also reveals the inadequacy of automatic evaluation metrics for capturing tutoring quality. Not only are the existing metrics unable to capture faithfulness to the learning material and the student dialog history, but they also cannot capture moves of good human tutors that allow learners the space for reflection, explanation, follow-ups, and real engagement in the process of learning.

Based on our findings, we end with an outline of potential avenues of future research (\cref{sec:recommend}). %
We 
hope that our paper will bring attention to this underexplored natural language processing application with the potential for significant social good.

\section{The Dialog Tutoring Task}\label{sec:dialogtask}

Dialog tutoring can be described as a multi-turn interaction between two interlocutors, where 
one performs the role of a \emph{teacher} seeking to teach the other interlocutor who acts in the role of a \emph{student}.
We then can describe a dialog tutoring session formally as a sequence of turns $\mathcal{H} = (u_1, \dots, u_{|\mathcal{H}|})$ that are taken by either of the interlocutors. 
Each turn $u_t \in \mathcal{V}^\ast$ is a finite sequence of tokens from a %
vocabulary $\mathcal{V}$.

Further, each turn $u_t$ can be associated with a sequence of dialog acts $\mathbf{a}_t \in \mathcal{A}$
that indicate the action taken by the interlocutor in the corresponding turn. The dialog act is a key aspect of dialog tutoring as it can refer to the teaching strategy employed by the tutor. These may include strategies such as \textit{providing a hint} or \textit{seeking a clarification} (see Appendix \ref{ap:ped-strategies-da} for more details).
The set of dialog acts $\mathcal{A}$ is usually fixed according to a predefined taxonomy and may be split into two subsets $\mathcal{A} = \mathcal{A}_{\text{teacher}} \cup \mathcal{A}_{\text{student}}$, each corresponding to the teacher and student role.
Each dialog session $\mathcal{H}$ may also be accompanied with some \emph{grounding} information $K$, which grounds the response in relevant information and may refer to the teaching material that needs to be taught to the student.
This information $K$ may come in various formats, including images and videos. However, we restrict ourselves to using only text-based grounding in this work such that $K \in \mathcal{K} \subseteq \mathcal{V}^\ast$ is again a sequence of tokens from the common vocabulary $\mathcal{V}$ and $\mathcal{K}$ is used to describe the set of possible groundings (e.g., a textbook with a set of chapters). In Section \ref{methods} we derive different methods to model the role of the teacher, to which we restrict this work.

\subsection{Existing tutoring datasets}\label{sec:existing-dataset}

\begin{figure}[]
    \centering
    \includegraphics[width=\linewidth]{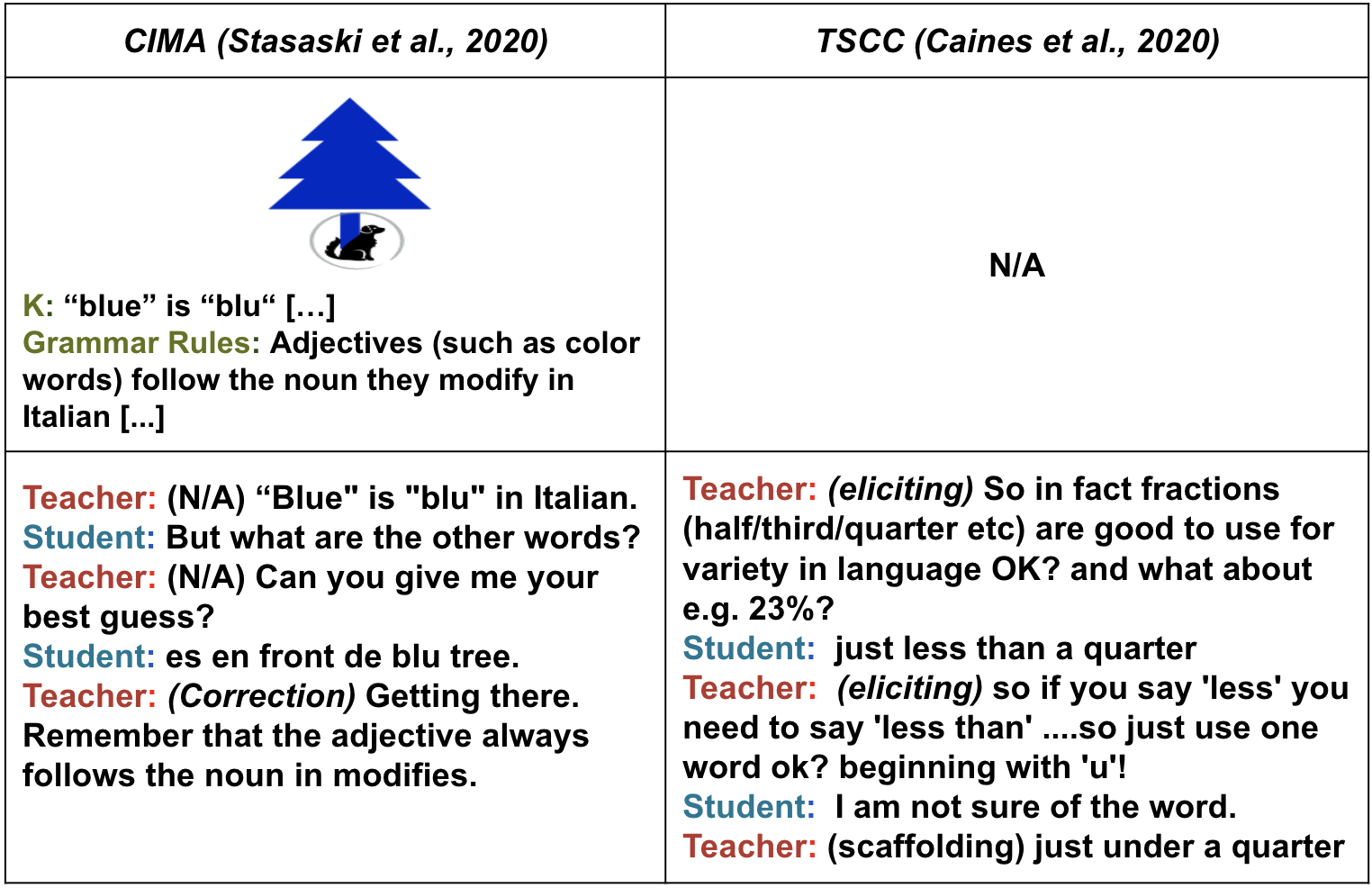}
    \caption{Examples of tutoring conversations from both datasets. The (image) grounding is shown in the second row and dialog acts in brackets indicate the pedagogical strategy. 
    }
    \label{fig:dialogue-tutoring-dataset-examples}
\end{figure}
To our knowledge, only three conversational tutoring datasets are openly available:
\textit{CIMA} \cite{stasaski-etal-2020-cima} is a crowd-sourced dataset, where annotators were asked to role-play students and teachers by working through an exercise on translating a prepositional phrase from
English to Italian, given an image and a shared set of concepts.
\textit{TSCC} \cite{caines2020teacher} uses real teachers leading one-on-one language tutoring sessions in English language learning, thus creating a more open-ended scenario.
Finally, \textit{TalkMoves} \cite{suresh2022talkmoves}. is a collection of scraped classroom transcripts of K-12 mathematics lesson videos that contain challenging, multi-party interactions.

\begin{table*}
    \centering
    \resizebox{\linewidth}{!}{
    \begin{tabular}{|l|cccccccc|}
    \hline
    Dataset        & Train samples & \#DA & Tgt. length & Src. length & \#prev. turns & corpus-div. & Flesch score & F1$(\hat{\mathbf{y}}, K)$ \\ \hline \hline
    CIMA           & 2,715   & 5 & 14.71 & 9.70 & 4.55 & 0.149 & 84.64 & 0.196\\
    TSCC           & 5,845   & 23 & 16.09 & 11.72 & 68.28 & 0.327 & 73.00 & -\\ \hline
    MultiWoZ 2.1   & 56,781  & 34$^\ast$ & 19.86 & 14.49 & 7.86 & 0.069 & 90.90 & -\\
    Schema-Guided Dialog & 164,982 & 10$^\ast$ & 14.30 & 10.36 & 11.38 & 0.049 & 95.37 & - \\
    DSTC9          & 19,184  & - & 21.61 & 11.65 & 11.70 & 0.050 & 81.85 & 0.47\\\hline
    Personachat    & 127,162 & - & 12.26 & 11.65 & 6.51 & 0.162 & 91.80 & 0.10\\
    FaithDial      & 18,357  & - & 21.72 & 17.33 & 4.54 & 0.274 & 83.28 & 0.47\\
    CMU\_DoG       & 81,468  & - & 14.49 & 18.23 & 18.73 & 0.178 & 79.54 & 0.02\\ \hline
    \end{tabular}
    }
    \caption{Dialogue dataset statistics. Target length and source length in avg. number of tokens (Bart tokenizer). \# prev. turns is avg. for each teacher response. Corpus-div is ngram entropy averaged for uni to four-grams. hlines separate: tutoring, task-oriented, open-domain dialog. $\ast$We only count system dialog acts.}\label{tab:dataset-stats}
\end{table*}

The scarcity of tutoring datasets stands in contrast to other dialog scenarios, where plenty of datasets have been proposed and studied.
For example, task-oriented dialog has been studied in domains like reservations \cite{wen2017camrest, budzianowski2018multiwoz, kim2020beyond} or public service information \cite{feng2020doc2diall}.
On the other hand, chit-chat or open-domain dialog has been studied on movies \cite{cmu_dog_emnlp18}, Wikipedia knowledge \cite{dinan2019wizard}, agent persona \cite{dinan2020second}, knowledge graphs \cite{Moon2019opendialkg}, and open-ended settings \cite{komeili-etal-2022-internet}.

Furthermore, we note the following limitations and characteristics of tutoring datasets, also in comparison to other dialog domains:
1) Low pedagogical quality (CIMA), 
2) Limited teaching strategies (all), 
3) Exclusive focus on classroom settings (TalkMoves), 
4) Small dataset size (all). 
5) Significantly larger context sizes (TSCC)
6) Harder readability according to the Flesch score (TSCC). We provide more evidence in Table \ref{tab:dataset-stats} which shows a comparison of dialog tutoring datasets with widely-used task-oriented and open-domain datasets.

\subsection{Related work on generative dialog models}
Similarly, while the advent of large pretrained models has sparked ample research on generative models for dialog \cite{bao2020plato, peng2020soloist, roller2020recipes, shuster2022blenderbot, lamda2022}, this has not carried over to research on tutoring systems, where existing solutions are predominantly rule-based and do not generate open-ended responses.
For example, the authors on CIMA define heuristics to select responses \cite{stasaski-etal-2020-cima}.
Pretrained transformers in general have only very recently been studied in this setting, however only for dialog act classification \cite{suresh-etal-2022-fine} and to study the pedagogical ability of existing large pretrained models \cite{tack_ai_2022}.

\section{Dialog Tutoring Models}
\label{methods}
After introducing the dialog tutoring task, this section highlights the models we evaluate on the task. We note that our aim is an analysis of existing models.

We explore turn-level models that can generate a teacher response $\mathbf{y} \coloneqq u_{t+1}$ given a tutoring session $\mathcal{H} = (u_1, \dots, u_{|\mathcal{H}|})$. %
During training, we obtain the dialog history by teacher forcing, i.e., we take the ground-truth dialog history.
Furthermore, we do not model the problem of retrieving grounding information but rather assume it as given.

\paragraph{Generative Model}

In order to study if generative models can capture a \emph{given} teaching strategy, we first derive a model that assumes the ground-truth dialog act sequence $\mathbf{a} = \{\mathbf{a}_1, \dots, \mathbf{a}_{|\mathcal{H}|}\}$ to be given as an input.
Then, given dialog history $\mathcal{H}_{< t} = \{u_1, \dots, u_t \}$, grounding information $K$ and $\mathbf{a}_{t+1} \subseteq \mathcal{A}_{\text{teacher}}$, the set of dialog acts relevant at timestep $t+1$, the teacher response $\mathbf{y}$ is generated according to a locally-normalized language generation model. 
In the case that no grounding information $K$ is given, the dependency on $K$ may be dropped.
\begin{eqnarray}
    \label{eq:grounded_generation}
        \mathbf{y}^\star &=& \argmax_{\mathbf{y} \in \mathcal{V}^\ast} \{ p\left(\mathbf{y} \mid \mathbf{a}_{t+1}, \mathcal{H}_{< t}, K\right) \}\\\nonumber
        &=& \argmax_{\mathbf{y} \in \mathcal{V}^\ast} \prod_{i=1}^{|{\bf y}|} \{ p\left(y_i \mid \mathbf{y}_{< i}, \mathbf{a}_{t+1}, \mathcal{H}_{<t}, K\right) \} %
\end{eqnarray}

We separate the turns in the dialog by special $\langle \text{teacher} \rangle$ and $\langle \text{student} \rangle$ tags and prepend the dialog act as a special token, followed by a special $\langle \text{knowledge} \rangle$ token and the grounding information $K$ as the input to the encoder.
In CIMA we encode the triples defining the grounding information in a simple natural language format, where we separate the English and Italian words for an object, color, and preposition as well as the whole phrase by the word "is", for example as "blue is blu" in Figure \ref{fig:dialogue-tutoring-dataset-examples}.
Further, we add the grammar rules separated by a special token. 
We study different models to parametrize $p$ that are described in Section \ref{sec:experiments_models}.

Finally, we use the version of {\bf CTRL} \cite{keskarCTRL2019} presented by \citet{rashkin2021ctrl}.
The aim of the model is to improve the faithfulness of grounded response generation models, a significant problem in neural language generation %
\cite{roller2020recipes} which holds high importance in the field of tutoring, where one trusts a teacher to present correct information. 
The model is augmented by a sequence of control tokens that are intended to steer the generations to desirable properties. We use the \emph{lexical overlap} and \emph{entailment tokens}, which we obtain as follows.
In training, the lexical overlap is measured on a token-level between ground-truth response and grounding.
Then, three equally sized buckets are created indicating low, medium, and high overlap which is indicated by a control token.
Entailment is determined by an MNLI model and again a corresponding token is added.
At test time, we always use the token that encourages the desirable property, in this case high lexical overlap and entailment.
Finally, using a sequence of control tokens $\mathbf{c}$, the model from equation \ref{eq:grounded_generation} becomes:
\vspace{-0.3em}
\begin{equation}
        \vspace{-0.3em}
        p\left(\mathbf{y} \mid \mathbf{a}_{t+1}, \mathbf{c},\mathcal{H}_{<t}, K \right)
    \label{eq:ctrl}
\end{equation}

\paragraph{Joint Model}
In order to study how well current neural models can decide on a reasonable teaching strategy and perform in real case scenarios, we also implement a model that first decides the dialog act $a_{t+1} \in \mathcal{A}_{\text{teacher}}$ (instead of assuming the ground-truth dialog act) and then uses it to generate a response ${\bf y} = u_{t+1}$.
We use a simple model that again takes the grounding and dialog context as input but now generates the concatenation of dialog act and response in one utterance, akin to SOLOIST \cite{peng2020soloist}.
Thus, for a given $\tilde{\mathbf{y}} \coloneqq \mathbf{a}_{t+1} \circ \mathbf{y}$ with act sequence $\mathbf{a}_{t+1}$ of length $N$ and response $\mathbf{y}$ of length T, the model is
\begin{align}
    \begin{split}
        p\left(\tilde{\mathbf{y}} \mid K, \mathcal{H}_{<t} \right)
        = \prod_{i=1}^{m+N} p\left(\tilde{y}_i \mid \tilde{\mathbf{y}}_{<i}, K, \mathcal{H}_{<t} \right)%
    \end{split}
\end{align}
In training, we use teacher forcing and prepend $\mathbf{a}_{t+1}$ to $\mathbf{y}$ to obtain the label sequence.
At test time, the model performs a beam search over the dialog act sequence and response jointly.

\paragraph{Retrieval-based model}
Since generative models are known to produce erroneous outputs that are factually incorrect and potentially inappropriate \cite{ji2022survey}, we also experiment with using a retrieval-based model that selects responses from the training corpus at test time.
As opposed to previous work on the topic (e.g., \citet{stasaski-etal-2020-cima}), we do not employ a rule-based model but rather a learned retrieval model that does not require handcrafting elaborate and possibly brittle rules.
Therefore, we use the {\bf Bi-Encoder} architecture \citep{mazare-etal-2018-training,dinan2019wizard} where a dialog context encoder $\text{enc}_{\mathcal{H}_{<t}; \theta}$ and a response encoder $\text{enc}_{\mathbf{y};\theta}$ encode context $\mathcal{H}_{<t}$ and possible responses $\mathbf{y}$ into a fixed size vector of same dimension $n$.
In our experiments, the weights $\theta$ of both encoders are shared.

The model is trained using contrastive learning.
Suppose we are given a training pair $\mathcal{H}, \hat{\mathbf{y}}$ from a training dataset $\mathcal{D}$ that we use for teacher forcing.
We then train the model by sampling a negative response $\bar{\mathbf{y}}$ from the set of responses in $\mathcal{D}$ and using the Triplet Loss criterion, which for a metric function $d : \mathbb{R}^n \times \mathbb{R}^n \rightarrow \mathbb{R}$ is defined as:
\begin{equation}
\begin{split}
    \mathcal{L}(\theta; \mathcal{H}, \hat{\mathbf{y}}, \bar{\mathbf{y}}) = [m &+ d(\text{enc}_{\mathcal{H}; \theta}(\mathcal{H}), \text{enc}_{\mathbf{y}; \theta}(\hat{\mathbf{y}})) \\
    &- d(\text{enc}_{\mathcal{H}; \theta}(\mathcal{H}), \text{enc}_{\mathbf{y}; \theta}(\bar{\mathbf{y}}))]_+ \text{,}
\end{split}
\end{equation}
where $m$ is a margin hyperparameter, and $d$ is the euclidean norm in our experiments.
Further, we do stratified sampling on CIMA to not select negatives from the same preposition, color, or object that might be false negatives. At test time, given a dialog context $\mathcal{H}_{<t}$, we choose a response $\mathbf{y}^\star$ from the training set $\mathcal{D}$ by maximum inner product search using the decision rule
\begin{equation}
    \mathbf{y}^\star = \argmax_{ \mathbf{y} \in \mathcal{D}} \{\text{enc}_{\mathcal{H}; \theta}(\mathcal{H}_{<t})^T \text{enc}_{\mathbf{y}; \theta}(\mathbf{y}) \} \text{.}
\end{equation}

\section{Experiments}

\begin{table*}
    \centering
    \resizebox{\textwidth}{!}{\begin{tabular}{|l|ccc|cc|}
    \hline
     & \multicolumn{3}{c}{\textbf{CIMA}} &  \multicolumn{2}{c|}{\textbf{TSCC}} \\
     Model & sBLEU / BLEU-1 ($\uparrow$) & BERT F1 ($\uparrow$) & $Q^2$ ($\uparrow$) & sBLEU / BLEU-1 ($\uparrow$) & BERT F1 ($\uparrow$) \\ \hline 
    Rule-based \citep{stasaski-etal-2020-cima}$^*$ & 0.34/- & 0.45 & - & - & - \\
    LSTM \citep{stasaski-etal-2020-cima}$^{*}$& 0.31/- & 0.53 & - & - & - \\     \hline
    Seq2seq  & 2.89 / 28.0 & 0.676 & 0.372 & - & - \\ %
    DialoGPT & 4.12 / 35.6 & 0.697 & 0.571 & 0.63 / 18.5 & 0.661 \\
    Bi-Encoder (RoBERTa-base)  & 5.89 / 23.9  & 0.690 & 0.344 & 1.367 / 8.8 & 0.638 \\
    CTRL (BART-base)    & 5.99 / 42.5  & 0.702  & 0.673 & - & -\\
    t5-small  & 7.36 / 34.0 & 0.672 & 0.676 & 2.72 / 12.1 & 0.646 \\
    BART-large & 8.61 / 38.7 & 0.715 & 0.673 & 1.85 / 13.7 & 0.658 \\
    BART-base & 9.58 / \textbf{42.5} & \textbf{0.726} & \textbf{0.680} & \textbf{2.67} / \textbf{18.6} & \textbf{0.670} \\
    mt5-small & \textbf{11.26} / 41.0 & 0.700 & 0.624 & 1.80 / 14.9 & 0.653 \\
 \hline
    BART-base$^\dagger$   & 5.61 / 41.03 & 0.707 & 0.642 & 1.90 / 15.4 & 0.659\\
    BART-large$^\dagger$  & 5.65 / 42.67 & 0.694 & 0.607 & 1.74 / 15.1 & 0.660 \\ \hline
    \end{tabular}}
    \caption{Comparison of models on CIMA and TSCC. We note that the strong sacrebleu differences are caused by the brevity penalty (all generative models generate too short sequences), $^\dagger$: use predicted dialog act label, others use ground-truth. * numbers taken from \cite{stasaski-etal-2020-cima} - here, numbers may not be comparable as there is no standard split in CIMA.}
    \label{tab:results1}
    \vspace{-1.0em}
\end{table*}

\label{sec:experiments_models}
We use the following models for parameterizing $p$ in Equation \ref{eq:ctrl}:
A {\bf sequence-to-sequence} model \cite{sutskever2014sequence} with a copy mechanism \cite{gu2016incorporating} trained from scratch.
A %
wide range of pretrained Transformers, namely {\bf BART} \cite{lewis2019bart}, {\bf DialoGPT} \cite{zhang2020dialogpt}, {\bf T5} \cite{raffel2020t5} and its multilingual version {\bf mT5} \cite{xue2021mt5}.

BART and T5 are pretrained encoder-decoder models that were trained on denoising and text-to-text tasks, respectively.
mT5 bases on T5 but is multilingual which might help with the code-switched utterances in CIMA.
Lastly, DialoGPT is an autoregressive language model %
based on GPT-2 \cite{radford2019language} that was pretrained on a large dialog dataset obtained from Reddit.
With this, we intend to study whether large-scale dialog-specific pretraining can aid in training educational tutors, as well.

\paragraph{Implementation Details}
We implement our experiments using the Huggingface transformers library and finetune the checkpoints provided as part of it for all Transformer-based models.
For these models, we use an initial learning rate of $3.25\times 10^{-5}$, 500 warmup steps and linear learning rate decay.
We train the models using a batch size of 8 and evaluate on the validation sets after each epoch.
In the end, we select the best model to be the one that has a minimal loss on the validation set.
The sequence-to-sequence baseline is trained from scratch using an initial learning rate of $0.001$ for 25,000 steps using the Adam optimizer and a dropout rate of $0.1$
We use beam search with a beam size of 10 to generate model responses.

\subsection{Dataset splits}
Since there are no official dataset splits for CIMA and TSCC, we split both datasets randomly into training, validation and test sets. We provide the exact split of the dataset in an accompanying code repository.
For CIMA, we use all such samples with less than three annotated tutor responses for training.
The other conversations are split randomly into equally-sized validation and test sets which results in 2715/300/300 samples each.

For TSCC, we split randomly along the conversations to obtain 82/10/11 training, validation, and test conversations each.
\subsection{Evaluation metrics}\label{sec:eval_metrics}
To evaluate our models, we use the BLEU implementation provided by the sacrebleu package (sBLEU) \cite{sacre-bleu} to measure lexical overlap between generated and ground-truth response.
Furthermore, we use BERT F1 (BERTScore) to measure their semantic similarity.
Lastly, for CIMA we also calculate $Q^2$ \citep{honovich-etal-2021-q2} which measures the factual consistency of the response $\mathbf{y}$ with the grounding information $K$ by employing a question-answering based matching.
Both BERTScore and $Q^2$ have shown strong correlation with human judgements on factual consistency \citet{honovich-etal-2022-true-evaluating}.

\section{Results}

In this section, we summarize our main findings in terms of automatic evaluation.
First, we give an overview of the performance of different models that we train on CIMA and TSCC in Section \ref{comparison-section}.
Then, we assess their ability to stay faithful to teaching strategies (Section \ref{faithfulness-teaching-strategies}) and study how grounding annotations can influence the faithfulness of neural dialog tutors (Section \ref{grounding-data-influence}), before studying their scaling behavior with dataset size and complexity (Section \ref{scaling-laws-section}) and their generalization capabilities (Section \ref{generalization-concepts-section}). We then finish with an assessment of using education-specific data for pretraining (Section \ref{edu-specific-pretraining-section}).

\subsection{Comparison of different models}
\label{comparison-section}
Table \ref{tab:results1} shows the key results from our experiments. %
First, all automatic metrics are \textit{significantly higher} on CIMA, which indicates that the models can fit CIMA much better than TSCC, with which current approaches still struggle. We further analyse this finding in Section \ref{faithfulness-teaching-strategies} and show that this is because TSCC has \textit{richer teaching strategies which are harder to model}.
Our comparison also suggests that finetuning large pretrained \textit{Transformer models generally gives better results than the rule-based and LSTM model} reported in \cite{stasaski-etal-2020-cima}, and our implemented retrieval and sequence-to-sequence baselines. This illustrates the potential of LLMs for dialog tutoring.

We also see a significant difference among different LLMs.
Dialog-specific pretraining of DialoGPT does not help and gives worse results than BART and T5, primarily because the model tends to generate short and generic responses more often.
Multilingual pretraining in mT5 improves over T5 only in some metrics, notably in BLEU and BERT F1 on CIMA but not in terms of $Q^2$.
Similarly, adding control tokens to BART does not improve $Q^2$ or other automatic metrics.
Surprisingly, using very large models %
actually degrades performance in our experiments.
Finally, the last two rows show results obtained with our joint model that does not use the ground-truth dialog act but predicts it together with the response sequence and still provides reasonable performance.

\subsection{How well can generative models capture teaching strategies?}
\label{faithfulness-teaching-strategies}

\begin{table}
    \centering
    \resizebox{\linewidth}{!}{\begin{tabular}{|l|ccccc|}
        \hline
         & \multicolumn{5}{c|}{Method}\\ \hline
         & GT & BART$_{\text{base}}$ & BART$_{\text{large}}$  & CTRL & Retrieval \\ \hline
         DA F1 & 78.3 & \textbf{81.0} & 70.1 & 63.0 & 43.1 \\ \hline
    \end{tabular}}
    \caption{F1 score of the dialog act classification based on the generated responses of our models.
    \label{tab:posteriori_da_prediction}}
\end{table}
\begin{figure}
\centering
  \resizebox{.75\columnwidth}{!}
  {\begin{tikzpicture}
    \begin{axis}[
        ybar,
        xmin=0,xmax=6,
        ymin=0, ymax=0.6,
        bar width=12,
        xtick={0,1,2,3,4,5,6},
        xticklabels={{},{Hint},{Question},{Correction},{Confirmation},{Other}, {}},
        xticklabel style={rotate=30},
        ]
        \addplot+[teal, opacity=0.5, draw=black] coordinates
        {
            (5, 0.0)
            (4, 0.052)
            (3, 0.147)
            (2, 0.276)
            (1, 0.526)
          };
        \addplot+[brown, opacity=0.75, draw=black] coordinates
        {
            (5, 0.037)
            (4, 0.063)
            (3, 0.228)
            (2, 0.237)
            (1, 0.434)
          };
          \end{axis}

\end{tikzpicture}}
\caption{Distribution of \textcolor{teal}{predicted} and \textcolor{brown}{ground-truth} dialog acts on CIMA. \label{fig:da_distribution_cima}}

\end{figure}
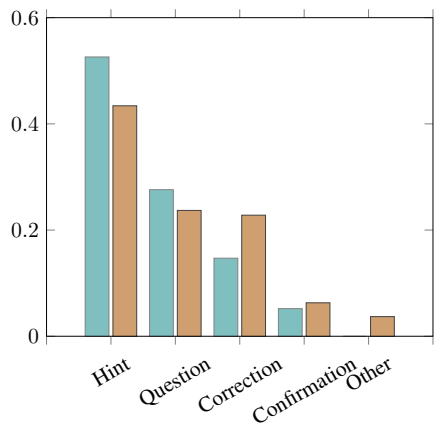
We study this question first by evaluating the dialog act prediction accuracy of our joint model.
We find that it is \textit{significantly low} on TSCC with $21.8$ compared to $71.2$ on CIMA for BART-base which indicates significant room for improvement.
Notably, the joint model tends to predict more frequently occuring dialog acts, which results in fewer follow-up questions and "Other" never being predicted in CIMA, the least frequent act in the data. The distribution of dialog acts in the ground-truth annotations and model predictions with a BART-base joint model is in Figure \ref{fig:da_distribution_cima}.

Then, we evaluate how well different models can stick to a given ground-truth dialog act by predicting the dialog act of the \emph{generated response} with a BART-base model trained to predict the ground-truth dialog act sequence based on the ground-truth response. 
The results are shown in Table \ref{tab:posteriori_da_prediction}.
Notably, \textit{BART-base performs better than the ground-truth annotations}. The CTRL model, on the other hand, has worse performance since the control tokens do not respect tutoring principles (e.g., lexical overlap to grounding discourages follow-up questions in favor of just giving hints).

\subsection{Does grounding in learning concepts help?}
\label{grounding-data-influence}
\begin{table}
    \centering
    \resizebox{\columnwidth}{!}{\begin{tabular}{|l|ccc|}
    \hline
    Model          & sBLEU ($\uparrow$) & BERT F1 ($\uparrow$) & $Q^2$ ($\uparrow$)\\ \hline \hline
    BART-base    & 6.69 / 38.6 & 0.718 & 0.571 \\ 
    $\;\;$ + triples     & 9.20 / \textbf{45.3} & \textbf{0.730} & 0.642 \\
    $\;\;\;$ + grammar rules & \textbf{9.58} / 42.5 & 0.726 & \textbf{0.680} \\
    \hline
    \end{tabular}}     \caption{Comparison of models with different inputs on CIMA. Triples are made up of preposition, object, and color translations. Grammar rules are a textual description of a learning concept.
    }
    \label{tab:results}
     \vspace{-1.0em}
\end{table}

Prior work has shown that grounding responses in relevant data can improve their quality, especially in terms of faithfulness \cite{shuster2021retrievalaugmentation}.
We intend to validate this for dialog tutoring by studying three models with different inputs on CIMA.
The first model is not provided grounding information, whereas
the second and third are grounded in learning concepts (cf. Equation \ref{eq:grounded_generation}) with one using only the (preposition, object, color) triples and the other making use of additional grammar rules.
The results with these models are shown in Table \ref{tab:results} and suggest that \textit{grounding responses in relevant knowledge helps the model to produce better and more faithful responses}.

\subsection{How do models scale with more data?}
\label{scaling-laws-section}
Due to the limited availability of high-quality pedagogical datasets and the time-consuming process of authoring new materials \citep{maclellan2020domain}, it is important to understand how quickly generative models can generalize to new settings.
\begin{figure}
\centering
  \begin{minipage}{.5\linewidth}
  \centering
 \resizebox{\columnwidth}{!}{\begin{tikzpicture}
    \begin{axis}[
      xmin=0,xmax=109,
      ymin=0,ymax=99,
      axis y line*=left,
      xlabel={\% of training data used},
      xlabel near ticks,
      ylabel={$Q^2 \cdot 100$ \ref{pgfplots:plot1}},
      ylabel near ticks
    ]
    \addplot+[olive,mark=*,mark options={fill=gray}, domain=35:0] coordinates {(50, 100)};
  \label{pgfplots:plot1}
    \addplot+[olive,mark=*,mark options={fill=gray}, domain=35:0] coordinates {
    (100, 68.0)
    (90, 63.2)
    (80, 61.6)
    (70, 63.4)
    (60, 64.1)
    (50, 56.5)
    (40, 55.9)
    (30, 55.3)
    (20, 48.4)
    (10, 48.1)
  };
    \end{axis}
    \begin{axis}[
      xmin = 0, xmax = 109,
      ymin = 0, ymax = 29,
      hide x axis,
      axis y line*=right,
      yticklabels={,,},
      ytick style={draw=none}
    ]
    \addplot+[cyan,mark=*,mark options={fill=gray}, domain=35:0] coordinates {(50, 100)};
    \label{pgfplots:plot2}
      \addplot+[cyan,mark=*,mark options={fill=gray}, domain=35:0] coordinates {
      (100, 9.59)
      (90, 8.92)
      (80, 9.89)
      (70, 8.67)
      (60, 8.61)
      (50, 9.10)
      (40, 8.69)
      (30, 8.99)
      (20, 7.36)
      (10, 3.11)
      };
    \end{axis}
  \end{tikzpicture}}
  \scriptsize (a)
  \end{minipage}%
  \begin{minipage}{.5\linewidth}
  \centering 
\resizebox{\columnwidth}{!}{\begin{tikzpicture}
    \begin{axis}[
      xmin=0,xmax=7,
      ymin=0,ymax=99,
      axis y line*=left,
      xlabel={No. of concepts},
      xlabel near ticks,
      yticklabels={,,},
      ytick style={draw=none}
    ]
    \addplot+[olive,mark=*,mark options={fill=gray}, domain=35:0] coordinates {(50, 100)};
  \label{pgfplots:plot1}
    \addplot+[olive,mark=*,mark options={fill=gray}, domain=35:0] coordinates {
      (1, 55.7)
      (2, 58.3)
      (3, 58.3)
      (4, 47.5)
      (5, 48.0)
      (6, 46.3)
  };
    \end{axis}
    \begin{axis}[
      xmin = 0, xmax = 7,
      ymin = 0, ymax = 29,
      hide x axis,
      axis y line*=right,
      ylabel={sBLEU \ref{pgfplots:plot2}},
      ylabel near ticks
    ]
    \addplot+[cyan,mark=*,mark options={fill=gray}, domain=35:0] coordinates {(50, 100)};
    \label{pgfplots:plot2}
      \addplot+[cyan,mark=*,mark options={fill=gray}, domain=35:0] coordinates {
      (1, 6.2)
      (2, 6.2)
      (3, 6.6)
      (4, 7.5)
      (5, 8.8)
      (6, 7.9)
      };
    \end{axis}
  \end{tikzpicture}}
  \scriptsize (b)
  \end{minipage}%
   \caption{Performance of BART-base on CIMA as a function of: (a) training data size uniformly sampled from the training data, (b) the number of concepts, where only the specific number of concepts is retained and all others are excluded.
   \label{tab:cima_figures}}
   \vspace{-1.0em}
\end{figure}
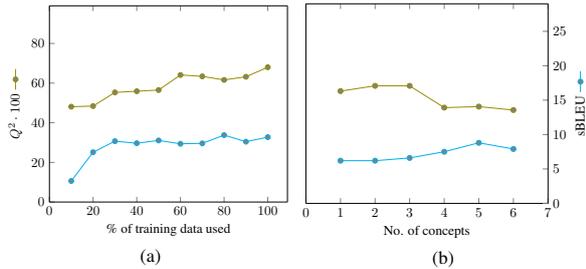
Thus, we assess how well the model can model tutoring in low-resource scenarios. We construct a study, where we randomly sample subsets of the CIMA training set and test the performance of the various models.
We can see from Figure \ref{tab:cima_figures}(a) that with more training data, the faithfulness of responses appears to improve and is not saturated before we reach the full training set.
This supports the intuition that \textit{additional training data might improve the performance further}.

Similarly, we study how well our model can deal with an increase in complexity with respect to learning concepts at similar training data sizes.
Therefore, we construct different training datasets with 735 samples and a varying number of concepts each time.
We begin by taking samples concerned with the concept ``in front of the'' and evaluate exclusively on it, gradually adding new concepts.
Figure \ref{tab:cima_figures}(b) suggests that $Q^2$ drops sharply at four concepts.
BLEU on the other hand increases, and this might be due to the metric encouraging generic utterances that, for example, repeat a grammar rule.

\subsection{Can models generalize to new concepts?}
\label{generalization-concepts-section}
As the students progress and gain new knowledge, it might be a desirable property of dialog tutoring models to be able to handle new concepts that suit this increase in prior knowledge.
Hence, we study how well our CIMA model can generalize to new concepts that it has not seen in training, for example, a new preposition.
For this analysis, we create a set-up where we first train the model on all of the training data and evaluate on the subset of samples for each preposition separately.
We then compare this number to a model that is not trained on the corresponding concept it is evaluated on, creating a zero-shot set-up which we carry out for a grounded and ungrounded response generation model.
As measured by $Q^2$ (cf. Table \ref{tab:results-learning-concepts}), this model can indeed \textit{generalize to new concepts well, albeit with performance degradation}.
Furthermore, \textit{grounding information improves generalization} as these define the learning concept (in this case the preposition) and how it is used.
Without this information, we observe that the model generates generic responses more often.

\begin{table}
\centering
\resizebox{\columnwidth}{!}{\begin{tabular}{|l|l|c|c|c|}
\hline
Concept & \#Samples & full data & zero-shot & zero-shot \\
& & & &  without grounding\\\hline
& train/test & $Q^2$ & $Q^2$ & $Q^2$\\
is behind the & 549/90 & 0.698 & 0.603 & 0.533 \\
is in front of the & 735/84 & 0.616 & 0.512 & 0.500\\
is next to the & 547/51 & 0.497 & 0.539 & 0.483\\
is on top of the & 224/30 & 0.683 & 0.578 & 0.567\\
is under the & 270/24 & 0.854 & 0.646 & 0.625\\
is inside of the & 390/21 & 0.579 & 0.643 & 0.190\\\hline
all concepts & 2715 / 300 & 0.644 & 0.570 & 0.502\\ \hline
\end{tabular}}
\caption{Performance of a grounded BART-base model by learning concept. Full data uses the entire training data and zero-shot removes the concept of the row from the training data.\label{tab:leave_one_out}}
\label{tab:results-learning-concepts}
\vspace{-1.0em}
\end{table}

\subsection{Does education-specific pre-training help?}
\label{edu-specific-pretraining-section}
\begin{table}[t]
\small
\begin{center}
\begin{tabular}{|l|ccc|}
\hline
Method & sBLEU  & BERT F1 & $Q^2$ \\
\hline
BART-base & 6.69 / 38.6 & 0.718 & 0.571 \\
+ Ed. data & \textbf{7.31} / \textbf{41.4} & \textbf{0.727} & 0.577 \\
+ Non-Ed. data & 6.60 / 39.4 & 0.721 & \textbf{0.583} \\
\hline
\end{tabular}
\end{center}
\caption{\label{tab:edu_spec_pretrain} 
Influence of pretraining on educational and non-educational data. Please note that no grounding information is used in this setting.
}
\end{table}
As educational data are widely available on the internet, next we study how education-specific pre-training effect results. In Table \ref{tab:edu_spec_pretrain}, we show results obtained with finetuning a BART-base model directly on CIMA and pretraining it on tutoring dialogs from TSCC or non-tutoring dialogs from MultiWoZ 2.1 \cite{eric2019multiwoz}, Personachat \cite{zhang2018personalizing}, CMU DoG \cite{cmu_dog_emnlp18}, DSTC9 \cite{kim2020beyond} and Topicalchat \cite{Gopalakrishnan2019}.
In both cases, we only see \textit{minor improvements}, which may be explained by the different dataset settings and the lack of a unified dialog act taxonomy.

\section{Human Evaluation}
We further evaluate previously assessed models with human judgments firstly by obtaining quality estimates according to different criteria and secondly by conducting a simulation study, where expert annotators are asked to provide novel rewritings of existing conversations and to categorize errors made by the model.

\subsection{Quality of the generated responses}
\label{human-quality}
\begin{figure}[h!]
    \centering
    \includegraphics[width=\linewidth]{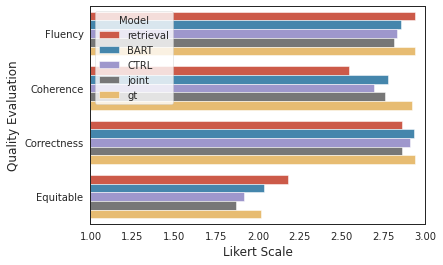}
    \caption{ Comparison of models on four criteria (reporting $M$) in the human quality evaluation. We observed high $SD$ for coherence and equitable metrics.
    }
    \label{fig:dialogue-tutoring-intro}
\end{figure}

We perform a human quality evaluation of the generated response for four models - retrieval (Bi-Encoder), BART-base, BART-base$_{\text{CTRL}}$ and the joint model (BART-base). A randomly chosen subset of the CIMA test set conversations were annotated by 4 annotators (with one annotator speaking C1 level Italian). All annotators labeled 60 examples in total, of which 20 overlapped. To further distinguish the quality of training data for the models, we annotated ground-truth responses on a small sample of 20 examples. We evaluate the following criteria on a 3-point Likert scale (disagree to completely agree) and outline our findings in the following, as shown in Figure \ref{fig:dialogue-tutoring-intro}.
    \paragraph{\bf Fluency} \textit{"The response is grammatically correct and fluent."}
    We find that all models have very high fluency scores.
    \paragraph{\bf Coherence} \textit{"The response naturally follows up on previous utterance and context and has no logical conflicts with the context or DA label."}
    We find that all generative models are able to produce coherent responses but not the retrieval model.
    \paragraph{\bf Correctness} \textit{"The response is factually correct and respects learning concepts being taught."}
    All models score comparable to ground-truth responses on the constrained CIMA dataset.
    It is noteworthy, however, that a response may be correct in itself but not coherent with the context or the grounding (often the case in the retrieval model), and this could explain the discrepancy between correctness and our automatic $Q^2$ scores.
    \paragraph{\bf Equitable tutoring} \textit{"The response gives a learning opportunity for the student by providing space for reflection, explanation, pointing to follow-up challenge, or engaging student in other ways."}
    Here we find significant deficiencies not only for our evaluated models but notably also for the annotated ground-truth responses (gt).
    This might explain the insufficiencies in the responses as they reflect this distributional behavior of the training data.
    We think that future dataset collections should take better care of this property and resort to more expert annotators as opposed to crowdsourcing.

\begin{table}
\small
    \centering
    \begin{tabular}{|l|cc|}
        \hline
         Quality Attribute & sBLEU & BERTScore \\ \hline
        Fluency & 0.14 & 0.12 \\
        Coherence & 0.17 & 0.26 \\
        Correctness & 0.06 & 0.15 \\
        Equitable Tutoring & 0.08 & 0.16 \\ \hline
    \end{tabular}
    \caption{Pearson correlation coefficients between the human judgements on our quality criteria and automatic metrics.
    \label{tab:correlations}}
    \vspace{-1.0em}
\end{table}
Furthermore, Table \ref{tab:correlations} shows that our automatic metrics correlate poorly with human judgements.

\subsection{User study with a learning interface}
\label{user-study-section}
Lastly, we seek to study how well dialog tutoring models can perform in a realistic setting with questions obtained from real users (containing out-of-distribution samples) and not the fixed dataset.
Therefore, we randomly sampled conversations from the CIMA test set. We asked two C1-level expert Italian speakers to 1) rephrase these conversations using a conversational dialogue interface and 2) assign erroneous model responses to predefined error categories.
The interface used in the qualitative evaluation is shown in Figure \ref{fig:dialogue-tutoring-interface}.
We obtain all model responses from the BART-base model that first predicts the dialog act and then the response.
Error categories adopted from previous work \cite{bommasani2021opportunities} describe the ideal behavior of tutoring models as simulating the behavior of good human teachers along two dimensions: 
\paragraph{\bf Understanding} \textit{"Being able to understand and reason about student solutions, misconceptions, and learning concepts."}
We find that of the 20 modified conversations, 45\% exhibit \textit{Understanding errors}, such as an incorrect solution assessment or incorrect translations.
\paragraph{\bf Pedagogy}\textit{"Being able to use effective pedagogy to instruct students."}
We find that 10\% of the responses exhibit \textit{Pedagogical errors}, for example telling the correct solution directly without offering any engagement point to the student. 

50\% of the conversations were labeled good by the annotators.
Examples of the conversations are available in Table \ref{tab:user-study-questions}.

\section{Discussion: Towards More Equitable and Faithful Tutoring Systems}\label{sec:recommend}
In this section, we outline directions of research that we think can be important steps towards more equitable and faithful tutoring models.
Namely, we first address the small scale and quality of current tutoring datasets and cast doubt on the crowdsourcing data quality checks.
Then, we suggest ways of improving the underperformance of both equitable tutoring and teaching strategy prediction identified in current generative models under these constraints by drawing from learning sciences literature.
Finally, we outline desiderata for more reliable dialog evaluation of neural tutoring models.

\paragraph{Datasets} Based on the analysis in \cref{sec:existing-dataset} and Table \ref{tab:dataset-stats}, we think that the community will benefit from a dataset that lies between CIMA and TSCC in terms of its difficulty. 
Moreover, the low equitable tutoring scores of CIMA's ground-truth responses indicate that crowdsourcing with untrained annotators can lead to low pedagogical quality. 
A similar observation has been found by human evaluation for the TSCC dataset \cite{tack_ai_2022}. 
Finally, we encourage the establishment of better dialog act taxonomies that are backed by learning sciences research. 
As outlined in \cref{edu-specific-pretraining-section} and in \citet{he2022galaxy}, a unified taxonomy may also strongly aid in transfer learning.

\paragraph{Models} So far, dialog tutoring models have only covered limited domain-specific settings linked to a particular activity, such as learning Italian prepositions or solving math word problems. 
We argue that the community could benefit from working on problems common to learning in general, for example tracking problem-solving states and modeling pedagogies used by teachers. Here, knowledge tracing \citep{corbett1994knowledge} (the problem of estimating students’ skill mastery level) could be used for tracking problem-solving states and increasing the coherence of dialog tutoring conversations and dialog act selection performance which would contribute to better modeling of global teaching strategies.
Furthermore, validated instruction quality coding schemes \citep{michaels2010accountable,hennessy2016developing} used by classroom teachers can be computationally modeled \cite{demszky-etal-2021-measuring,ganesh-etal-2021-teacher} and incorporated into models.

We also think that recently proposed constrained decoding approaches that can balance between multiple criteria \citep{cold_decoding_2022} hold great promise in improving faithfulness in complex tutoring dialogs.
Finally, as data collection is labor-intensive in expert domains, we see great potential in few-shot learning methods, such as prompt-based methods \citep{schick2022true}.

\paragraph{Evaluation}

Our experiments highlight the insufficiencies of current automatic dialog evaluation metrics, as both BLEU and BertScore show comparatively low correlation with our collected human judgements from \cref{human-quality}.
This is in line with previous research \cite{mehri2020usr, mehri2022report} and shows the necessity not only for better automatic evaluation metrics but also for verification based on human judgements or user studies that should incorporate criteria relevant to tutoring (e.g., equitable tutoring outcomes).
Metrics that incorporate task success, which have been used in task-oriented dialog systems \citep{budzianowski2018multiwoz}, are a promising direction of future research for automatic evaluation.

\section{Conclusion}
In this work, we reflected on the state of research in dialog tutoring and explored the potential of neural generative models in this domain. We found some promising initial results with these models in comparison to rule- or retrieval-based methods.
However, we also established limitations of currently available benchmarks and evaluation criteria.
Furthermore, we showed that there are a number of challenges that need to be addressed before neural generative models of text can be deployed as intelligent tutoring systems on a larger scale, such as controllability and being able to model a sound pedagogical strategy.
Based on these findings, we outline potential avenues for future research.

\section*{Limitations}
A key limitation of our work is the use of only two available tutoring datasets. Despite a limited number of datasets available in this domain, using the TalkMoves dataset \citep{suresh2022talkmoves} could help further generalize our findings. This remains an avenue for future work. 

Based on the prior work, we focused on the specific conversational goal of dialog tutors which is providing learning aid for students' skill development and more opportunities to learn. While this is the most widespread type \cite{wollny2021we}, it is not covering all the goals of human tutors, and other aspects could be important, for example, rapport-building or mentoring on the meta-cognitive level. We acknowledge this both as a prerequisite for our work and at the same time as a limitation. For further discussion we refer the reader to Appendix \ref{sec:equitable} and \ref{appendix-conversational-goals}. 

Finally, our user study could be further extended with more participants. In the future, we plan a more comprehensive study with real language learners using an end-to-end dialog tutoring system.

\section*{Ethics Statement}
We do not foresee any significant harm directly as a result of our work. Having said that, we must understand that automatic tutoring is a high-stake setting that can pose significant harm if appropriate care is not taken before the deployment of these systems. Issues of biases and lack of trust, and other ethical issues such as privacy concerns must be considered. Considering learners only as data points within a neural dialog tutoring context may prevent us from seeing the societal and socioeconomic barriers that they may be up against, thereby running the risk of not only failing to help relevant learner subgroups but also sometimes giving additional privileges to those who use these systems.

\section{Acknowledgements}
This project was made possible by an ETH AI Center Doctoral Fellowship to Jakub Macina with partial support by the Asuera Stiftung and the ETH Zurich Foundation and has received funding by the German Federal Ministry of Education and Research and the Hessian Ministry of Higher Education, Research, Science and the Arts within their joint support of the National Research Center for Applied Cybersecurity ATHENE.
We thank the group members and our reviewers for their valuable feedback.

\bibliography{anthology,custom_new}

\begin{thebibliography}{66}
\expandafter\ifx\csname natexlab\endcsname\relax\def\natexlab#1{#1}\fi

\bibitem[{Adiwardana et~al.(2020)Adiwardana, Luong, So, Hall, Fiedel,
  Thoppilan, Yang, Kulshreshtha, Nemade, Lu et~al.}]{adiwardana2020towards}
Daniel Adiwardana, Minh-Thang Luong, David~R So, Jamie Hall, Noah Fiedel, Romal
  Thoppilan, Zi~Yang, Apoorv Kulshreshtha, Gaurav Nemade, Yifeng Lu, et~al.
  2020.
\newblock \href {https://arxiv.org/abs/2001.09977} {Towards a human-like
  open-domain chatbot}.
\newblock \emph{ArXiv preprint}, abs/2001.09977.

\bibitem[{Bao et~al.(2021)Bao, He, Wang, Wu, Wang, Wu, Guo, Liu, and
  Xu}]{bao2020plato}
Siqi Bao, Huang He, Fan Wang, Hua Wu, Haifeng Wang, Wenquan Wu, Zhen Guo,
  Zhibin Liu, and Xinchao Xu. 2021.
\newblock \href {https://doi.org/10.18653/v1/2021.findings-acl.222} {{PLATO-2}:
  Towards building an open-domain chatbot via curriculum learning}.
\newblock In \emph{Findings of the Association for Computational Linguistics:
  ACL-IJCNLP 2021}, pages 2513--2525, Online. Association for Computational
  Linguistics.

\bibitem[{Bommasani et~al.(2021)Bommasani, Hudson, Adeli, Altman, Arora, von
  Arx, Bernstein, Bohg, Bosselut, Brunskill
  et~al.}]{bommasani2021opportunities}
Rishi Bommasani, Drew~A Hudson, Ehsan Adeli, Russ Altman, Simran Arora, Sydney
  von Arx, Michael~S Bernstein, Jeannette Bohg, Antoine Bosselut, Emma
  Brunskill, et~al. 2021.
\newblock On the opportunities and risks of foundation models.
\newblock \emph{arXiv preprint arXiv:2108.07258}.

\bibitem[{Budzianowski et~al.(2018)Budzianowski, Wen, Tseng, Casanueva, Ultes,
  Ramadan, and Ga{\v{s}}i{\'c}}]{budzianowski2018multiwoz}
Pawe{\l} Budzianowski, Tsung-Hsien Wen, Bo-Hsiang Tseng, I{\~n}igo Casanueva,
  Stefan Ultes, Osman Ramadan, and Milica Ga{\v{s}}i{\'c}. 2018.
\newblock \href {https://doi.org/10.18653/v1/D18-1547} {{M}ulti{WOZ} - a
  large-scale multi-domain {W}izard-of-{O}z dataset for task-oriented dialogue
  modelling}.
\newblock In \emph{Proceedings of the 2018 Conference on Empirical Methods in
  Natural Language Processing}, pages 5016--5026, Brussels, Belgium.
  Association for Computational Linguistics.

\bibitem[{Caines et~al.(2020)Caines, Yannakoudakis, Edmondson, Allen,
  P{\'e}rez-Paredes, Byrne, and Buttery}]{caines2020teacher}
Andrew Caines, Helen Yannakoudakis, Helena Edmondson, Helen Allen, Pascual
  P{\'e}rez-Paredes, Bill Byrne, and Paula Buttery. 2020.
\newblock \href {https://aclanthology.org/2020.nlp4call-1.2} {The
  teacher-student chatroom corpus}.
\newblock In \emph{Proceedings of the 9th Workshop on NLP for Computer Assisted
  Language Learning}, pages 10--20, Gothenburg, Sweden. LiU Electronic Press.

\bibitem[{Chi et~al.(1994)Chi, De~Leeuw, Chiu, and
  LaVancher}]{chi1994eliciting}
Michelene~TH Chi, Nicholas De~Leeuw, Mei-Hung Chiu, and Christian LaVancher.
  1994.
\newblock Eliciting self-explanations improves understanding.
\newblock \emph{Cognitive science}, 18(3):439--477.

\bibitem[{Chi and Wylie(2014)}]{chi2014icap}
Michelene~TH Chi and Ruth Wylie. 2014.
\newblock The icap framework: Linking cognitive engagement to active learning
  outcomes.
\newblock \emph{Educational psychologist}, 49(4):219--243.

\bibitem[{Cohen et~al.(2022)Cohen, Roberts, Molina, Butryna, Jin, Kulshreshtha,
  Hutchinson, Zevenbergen, Aguera-Arcas, ching Chang, Cui, Du, Adiwardana,
  Chen, Lepikhin, Chi, Hoffman-John, Cheng, Lee, Krivokon, Qin, Hall, Fenton,
  Soraker, Meier-Hellstern, Olson, Aroyo, Bosma, Pickett, Menegali, Croak,
  Díaz, Lamm, Krikun, Morris, Shazeer, Le, Bernstein, Rajakumar, Kurzweil,
  Thoppilan, Zheng, Bos, Duke, Doshi, Prabhakaran, Rusch, Li, Huang, Zhou, Xu,
  and Chen}]{lamda2022}
Aaron~Daniel Cohen, Adam Roberts, Alejandra Molina, Alena Butryna, Alicia Jin,
  Apoorv Kulshreshtha, Ben Hutchinson, Ben Zevenbergen, Blaise~Hilary
  Aguera-Arcas, Chung ching Chang, Claire Cui, Cosmo Du, Daniel De~Freitas
  Adiwardana, Dehao Chen, Dmitry~(Dima) Lepikhin, Ed~H. Chi, Erin Hoffman-John,
  Heng-Tze Cheng, Hongrae Lee, Igor Krivokon, James Qin, Jamie Hall, Joe
  Fenton, Johnny Soraker, Kathy Meier-Hellstern, Kristen Olson, Lora~Mois
  Aroyo, Maarten~Paul Bosma, Marc~Joseph Pickett, Marcelo~Amorim Menegali,
  Marian Croak, Mark Díaz, Matthew Lamm, Maxim Krikun, Meredith~Ringel Morris,
  Noam Shazeer, Quoc~V. Le, Rachel Bernstein, Ravi Rajakumar, Ray Kurzweil,
  Romal Thoppilan, Steven Zheng, Taylor Bos, Toju Duke, Tulsee Doshi,
  Vinodkumar Prabhakaran, Will Rusch, YaGuang Li, Yanping Huang, Yanqi Zhou,
  Yuanzhong Xu, and Zhifeng Chen. 2022.
\newblock Lamda: Language models for dialog applications.
\newblock In \emph{arXiv}.

\bibitem[{Corbett and Anderson(1994)}]{corbett1994knowledge}
Albert~T Corbett and John~R Anderson. 1994.
\newblock Knowledge tracing: Modeling the acquisition of procedural knowledge.
\newblock \emph{User modeling and user-adapted interaction}, 4:253--278.

\bibitem[{Demszky et~al.(2021)Demszky, Liu, Mancenido, Cohen, Hill, Jurafsky,
  and Hashimoto}]{demszky-etal-2021-measuring}
Dorottya Demszky, Jing Liu, Zid Mancenido, Julie Cohen, Heather Hill, Dan
  Jurafsky, and Tatsunori Hashimoto. 2021.
\newblock \href {https://doi.org/10.18653/v1/2021.acl-long.130} {Measuring
  conversational uptake: A case study on student-teacher interactions}.
\newblock In \emph{Proceedings of the 59th Annual Meeting of the Association
  for Computational Linguistics and the 11th International Joint Conference on
  Natural Language Processing (Volume 1: Long Papers)}, pages 1638--1653,
  Online. Association for Computational Linguistics.

\bibitem[{Dinan et~al.(2020)Dinan, Logacheva, Malykh, Miller, Shuster, Urbanek,
  Kiela, Szlam, Serban, Lowe et~al.}]{dinan2020second}
Emily Dinan, Varvara Logacheva, Valentin Malykh, Alexander Miller, Kurt
  Shuster, Jack Urbanek, Douwe Kiela, Arthur Szlam, Iulian Serban, Ryan Lowe,
  et~al. 2020.
\newblock The second conversational intelligence challenge (convai2).
\newblock In \emph{The NeurIPS'18 Competition}, pages 187--208. Springer.

\bibitem[{Dinan et~al.(2019)Dinan, Roller, Shuster, Fan, Auli, and
  Weston}]{dinan2019wizard}
Emily Dinan, Stephen Roller, Kurt Shuster, Angela Fan, Michael Auli, and Jason
  Weston. 2019.
\newblock \href {https://openreview.net/forum?id=r1l73iRqKm} {Wizard of
  wikipedia: Knowledge-powered conversational agents}.
\newblock In \emph{7th International Conference on Learning Representations,
  {ICLR} 2019, New Orleans, LA, USA, May 6-9, 2019}. OpenReview.net.

\bibitem[{Eric et~al.(2020)Eric, Goel, Paul, Sethi, Agarwal, Gao, Kumar, Goyal,
  Ku, and Hakkani-Tur}]{eric2019multiwoz}
Mihail Eric, Rahul Goel, Shachi Paul, Abhishek Sethi, Sanchit Agarwal, Shuyang
  Gao, Adarsh Kumar, Anuj Goyal, Peter Ku, and Dilek Hakkani-Tur. 2020.
\newblock \href {https://aclanthology.org/2020.lrec-1.53} {{M}ulti{WOZ} 2.1: A
  consolidated multi-domain dialogue dataset with state corrections and state
  tracking baselines}.
\newblock In \emph{Proceedings of the 12th Language Resources and Evaluation
  Conference}, pages 422--428, Marseille, France. European Language Resources
  Association.

\bibitem[{Feng et~al.(2020)Feng, Wan, Gunasekara, Patel, Joshi, and
  Lastras}]{feng2020doc2diall}
Song Feng, Hui Wan, Chulaka Gunasekara, Siva Patel, Sachindra Joshi, and Luis
  Lastras. 2020.
\newblock \href {https://doi.org/10.18653/v1/2020.emnlp-main.652} {doc2dial: A
  goal-oriented document-grounded dialogue dataset}.
\newblock In \emph{Proceedings of the 2020 Conference on Empirical Methods in
  Natural Language Processing (EMNLP)}, pages 8118--8128, Online. Association
  for Computational Linguistics.

\bibitem[{Freeman et~al.(2014)Freeman, Eddy, McDonough, Smith, Okoroafor,
  Jordt, and Wenderoth}]{freeman2014active}
Scott Freeman, Sarah~L Eddy, Miles McDonough, Michelle~K Smith, Nnadozie
  Okoroafor, Hannah Jordt, and Mary~Pat Wenderoth. 2014.
\newblock Active learning increases student performance in science,
  engineering, and mathematics.
\newblock \emph{Proceedings of the national academy of sciences},
  111(23):8410--8415.

\bibitem[{Ganesh et~al.(2021)Ganesh, Palmer, and
  Kann}]{ganesh-etal-2021-teacher}
Ananya Ganesh, Martha Palmer, and Katharina Kann. 2021.
\newblock \href {https://doi.org/10.18653/v1/2021.findings-acl.418} {What would
  a teacher do? {P}redicting future talk moves}.
\newblock In \emph{Findings of the Association for Computational Linguistics:
  ACL-IJCNLP 2021}, pages 4739--4751, Online. Association for Computational
  Linguistics.

\bibitem[{Gopalakrishnan et~al.(2019)Gopalakrishnan, Hedayatnia, Chen,
  Gottardi, Kwatra, Venkatesh, Gabriel, and Hakkani-Tür}]{Gopalakrishnan2019}
Karthik Gopalakrishnan, Behnam Hedayatnia, Qinlang Chen, Anna Gottardi, Sanjeev
  Kwatra, Anu Venkatesh, Raefer Gabriel, and Dilek Hakkani-Tür. 2019.
\newblock \href {https://doi.org/10.21437/Interspeech.2019-3079}
  {{Topical-Chat: Towards Knowledge-Grounded Open-Domain Conversations}}.
\newblock In \emph{Proc. Interspeech 2019}, pages 1891--1895.

\bibitem[{Graesser(2016)}]{graesser2016conversations}
Arthur~C Graesser. 2016.
\newblock Conversations with autotutor help students learn.
\newblock \emph{International Journal of Artificial Intelligence in Education},
  26(1):124--132.

\bibitem[{Graesser et~al.(2009)Graesser, D’MELLO, and
  Person}]{graesser2009meta}
Arthur~C Graesser, SIDNEY D’MELLO, and Natalie Person. 2009.
\newblock Meta-knowledge in tutoring.
\newblock In \emph{Handbook of metacognition in education}, pages 373--394.
  Routledge.

\bibitem[{Graesser et~al.(1995)Graesser, Person, and
  Magliano}]{graesser1995collaborative}
Arthur~C Graesser, Natalie~K Person, and Joseph~P Magliano. 1995.
\newblock Collaborative dialogue patterns in naturalistic one-to-one tutoring.
\newblock \emph{Applied cognitive psychology}, 9(6):495--522.

\bibitem[{Gu et~al.(2016)Gu, Lu, Li, and Li}]{gu2016incorporating}
Jiatao Gu, Zhengdong Lu, Hang Li, and Victor~O.K. Li. 2016.
\newblock \href {https://doi.org/10.18653/v1/P16-1154} {Incorporating copying
  mechanism in sequence-to-sequence learning}.
\newblock In \emph{Proceedings of the 54th Annual Meeting of the Association
  for Computational Linguistics (Volume 1: Long Papers)}, pages 1631--1640,
  Berlin, Germany. Association for Computational Linguistics.

\bibitem[{He et~al.(2022)He, Dai, Zheng, Wu, Cao, Liu, Jiang, Yang, Huang, Si
  et~al.}]{he2022galaxy}
Wanwei He, Yinpei Dai, Yinhe Zheng, Yuchuan Wu, Zheng Cao, Dermot Liu, Peng
  Jiang, Min Yang, Fei Huang, Luo Si, et~al. 2022.
\newblock Galaxy: A generative pre-trained model for task-oriented dialog with
  semi-supervised learning and explicit policy injection.
\newblock \emph{Proceedings of the AAAI Conference on Artificial Intelligence}.

\bibitem[{Hennessy et~al.(2016)Hennessy, Rojas-Drummond, Higham, M{\'a}rquez,
  Maine, R{\'\i}os, Garc{\'\i}a-Carri{\'o}n, Torreblanca, and
  Barrera}]{hennessy2016developing}
Sara Hennessy, Sylvia Rojas-Drummond, Rupert Higham, Ana~Mar{\'\i}a
  M{\'a}rquez, Fiona Maine, Rosa~Mar{\'\i}a R{\'\i}os, Roc{\'\i}o
  Garc{\'\i}a-Carri{\'o}n, Omar Torreblanca, and Mar{\'\i}a~Jos{\'e} Barrera.
  2016.
\newblock Developing a coding scheme for analysing classroom dialogue across
  educational contexts.
\newblock \emph{Learning, culture and social interaction}, 9:16--44.

\bibitem[{Honovich et~al.(2022)Honovich, Aharoni, Herzig, Taitelbaum,
  Kukliansy, Cohen, Scialom, Szpektor, Hassidim, and
  Matias}]{honovich-etal-2022-true-evaluating}
Or~Honovich, Roee Aharoni, Jonathan Herzig, Hagai Taitelbaum, Doron Kukliansy,
  Vered Cohen, Thomas Scialom, Idan Szpektor, Avinatan Hassidim, and Yossi
  Matias. 2022.
\newblock \href {https://doi.org/10.18653/v1/2022.naacl-main.287} {{TRUE}:
  Re-evaluating factual consistency evaluation}.
\newblock In \emph{Proceedings of the 2022 Conference of the North American
  Chapter of the Association for Computational Linguistics: Human Language
  Technologies}, pages 3905--3920, Seattle, United States. Association for
  Computational Linguistics.

\bibitem[{Honovich et~al.(2021)Honovich, Choshen, Aharoni, Neeman, Szpektor,
  and Abend}]{honovich-etal-2021-q2}
Or~Honovich, Leshem Choshen, Roee Aharoni, Ella Neeman, Idan Szpektor, and Omri
  Abend. 2021.
\newblock \href {https://doi.org/10.18653/v1/2021.emnlp-main.619} {Q$^{2}$:
  {E}valuating factual consistency in knowledge-grounded dialogues via question
  generation and question answering}.
\newblock In \emph{Proceedings of the 2021 Conference on Empirical Methods in
  Natural Language Processing}, pages 7856--7870, Online and Punta Cana,
  Dominican Republic. Association for Computational Linguistics.

\bibitem[{Ji et~al.(2022)Ji, Lee, Frieske, Yu, Su, Xu, Ishii, Bang, Madotto,
  and Fung}]{ji2022survey}
Ziwei Ji, Nayeon Lee, Rita Frieske, Tiezheng Yu, Dan Su, Yan Xu, Etsuko Ishii,
  Yejin Bang, Andrea Madotto, and Pascale Fung. 2022.
\newblock Survey of hallucination in natural language generation.
\newblock \emph{ACM Computing Surveys}.

\bibitem[{Keskar et~al.(2019)Keskar, McCann, Varshney, Xiong, and
  Socher}]{keskarCTRL2019}
Nitish~Shirish Keskar, Bryan McCann, Lav Varshney, Caiming Xiong, and Richard
  Socher. 2019.
\newblock \href {https://arxiv.org/abs/1909.05858} {{CTRL - A Conditional
  Transformer Language Model for Controllable Generation}}.
\newblock \emph{ArXiv preprint}, abs/1909.05858.

\bibitem[{Kim et~al.(2020)Kim, Eric, Gopalakrishnan, Hedayatnia, Liu, and
  Hakkani-Tur}]{kim2020beyond}
Seokhwan Kim, Mihail Eric, Karthik Gopalakrishnan, Behnam Hedayatnia, Yang Liu,
  and Dilek Hakkani-Tur. 2020.
\newblock \href {https://aclanthology.org/2020.sigdial-1.35} {Beyond domain
  {API}s: Task-oriented conversational modeling with unstructured knowledge
  access}.
\newblock In \emph{Proceedings of the 21th Annual Meeting of the Special
  Interest Group on Discourse and Dialogue}, pages 278--289, 1st virtual
  meeting. Association for Computational Linguistics.

\bibitem[{Komeili et~al.(2022)Komeili, Shuster, and
  Weston}]{komeili-etal-2022-internet}
Mojtaba Komeili, Kurt Shuster, and Jason Weston. 2022.
\newblock \href {https://doi.org/10.18653/v1/2022.acl-long.579}
  {{I}nternet-augmented dialogue generation}.
\newblock In \emph{Proceedings of the 60th Annual Meeting of the Association
  for Computational Linguistics (Volume 1: Long Papers)}, pages 8460--8478,
  Dublin, Ireland. Association for Computational Linguistics.

\bibitem[{Lewis et~al.(2020)Lewis, Liu, Goyal, Ghazvininejad, Mohamed, Levy,
  Stoyanov, and Zettlemoyer}]{lewis2019bart}
Mike Lewis, Yinhan Liu, Naman Goyal, Marjan Ghazvininejad, Abdelrahman Mohamed,
  Omer Levy, Veselin Stoyanov, and Luke Zettlemoyer. 2020.
\newblock \href {https://doi.org/10.18653/v1/2020.acl-main.703} {{BART}:
  Denoising sequence-to-sequence pre-training for natural language generation,
  translation, and comprehension}.
\newblock In \emph{Proceedings of the 58th Annual Meeting of the Association
  for Computational Linguistics}, pages 7871--7880, Online. Association for
  Computational Linguistics.

\bibitem[{Litman et~al.(2006)Litman, Ros{\'e}, Forbes-Riley, VanLehn, Bhembe,
  and Silliman}]{litman2006spoken}
Diane~J Litman, Carolyn~P Ros{\'e}, Kate Forbes-Riley, Kurt VanLehn, Dumisizwe
  Bhembe, and Scott Silliman. 2006.
\newblock Spoken versus typed human and computer dialogue tutoring.
\newblock \emph{International Journal of Artificial Intelligence in Education},
  16(2):145--170.

\bibitem[{MacLellan and Koedinger(2020)}]{maclellan2020domain}
Christopher~J MacLellan and Kenneth~R Koedinger. 2020.
\newblock Domain-general tutor authoring with apprentice learner models.
\newblock \emph{International Journal of Artificial Intelligence in Education},
  pages 1--42.

\bibitem[{Mazar{\'e} et~al.(2018)Mazar{\'e}, Humeau, Raison, and
  Bordes}]{mazare-etal-2018-training}
Pierre-Emmanuel Mazar{\'e}, Samuel Humeau, Martin Raison, and Antoine Bordes.
  2018.
\newblock \href {https://doi.org/10.18653/v1/D18-1298} {Training millions of
  personalized dialogue agents}.
\newblock In \emph{Proceedings of the 2018 Conference on Empirical Methods in
  Natural Language Processing}, pages 2775--2779, Brussels, Belgium.
  Association for Computational Linguistics.

\bibitem[{Mehri et~al.(2022)Mehri, Choi, D'Haro, Deriu, Eskenazi, Gasic,
  Georgila, Hakkani-Tur, Li, Rieser et~al.}]{mehri2022report}
Shikib Mehri, Jinho Choi, Luis~Fernando D'Haro, Jan Deriu, Maxine Eskenazi,
  Milica Gasic, Kallirroi Georgila, Dilek Hakkani-Tur, Zekang Li, Verena
  Rieser, et~al. 2022.
\newblock \href {https://arxiv.org/abs/2203.10012} {Report from the nsf future
  directions workshop on automatic evaluation of dialog: Research directions
  and challenges}.
\newblock \emph{ArXiv preprint}, abs/2203.10012.

\bibitem[{Mehri and Eskenazi(2020)}]{mehri2020usr}
Shikib Mehri and Maxine Eskenazi. 2020.
\newblock \href {https://doi.org/10.18653/v1/2020.acl-main.64} {{USR}: An
  unsupervised and reference free evaluation metric for dialog generation}.
\newblock In \emph{Proceedings of the 58th Annual Meeting of the Association
  for Computational Linguistics}, pages 681--707, Online. Association for
  Computational Linguistics.

\bibitem[{Michaels et~al.(2010)Michaels, O’Connor, Hall, and
  Resnick}]{michaels2010accountable}
Sarah Michaels, Mary~Catherine O’Connor, Megan~Williams Hall, and Lauren~B
  Resnick. 2010.
\newblock Accountable talk sourcebook: For classroom conversation that works.
\newblock \emph{Pittsburgh, PA: University of Pittsburgh Institute for
  Learning}.

\bibitem[{Miller et~al.(2017)Miller, Feng, Batra, Bordes, Fisch, Lu, Parikh,
  and Weston}]{miller2017parlai}
Alexander Miller, Will Feng, Dhruv Batra, Antoine Bordes, Adam Fisch, Jiasen
  Lu, Devi Parikh, and Jason Weston. 2017.
\newblock \href {https://doi.org/10.18653/v1/D17-2014} {{P}arl{AI}: A dialog
  research software platform}.
\newblock In \emph{Proceedings of the 2017 Conference on Empirical Methods in
  Natural Language Processing: System Demonstrations}, pages 79--84,
  Copenhagen, Denmark. Association for Computational Linguistics.

\bibitem[{Moon et~al.(2019)Moon, Shah, Kumar, and Subba}]{Moon2019opendialkg}
Seungwhan Moon, Pararth Shah, Anuj Kumar, and Rajen Subba. 2019.
\newblock \href {https://doi.org/10.18653/v1/P19-1081} {{O}pen{D}ial{KG}:
  Explainable conversational reasoning with attention-based walks over
  knowledge graphs}.
\newblock In \emph{Proceedings of the 57th Annual Meeting of the Association
  for Computational Linguistics}, pages 845--854, Florence, Italy. Association
  for Computational Linguistics.

\bibitem[{Moore et~al.(2004)Moore, Porayska-Pomsta, Varges, and
  Zinn}]{moore2004generating}
Johanna~D Moore, Kaska Porayska-Pomsta, Sebastian Varges, and Claus Zinn. 2004.
\newblock Generating tutorial feedback with affect.
\newblock In \emph{FLAIRS Conference}, pages 923--928.

\bibitem[{Nye et~al.(2014)Nye, Graesser, and Hu}]{nye2014autotutor}
Benjamin~D Nye, Arthur~C Graesser, and Xiangen Hu. 2014.
\newblock Autotutor and family: A review of 17 years of natural language
  tutoring.
\newblock \emph{International Journal of Artificial Intelligence in Education},
  24(4):427--469.

\bibitem[{Peng et~al.(2021)Peng, Li, Li, Shayandeh, Liden, and
  Gao}]{peng2020soloist}
Baolin Peng, Chunyuan Li, Jinchao Li, Shahin Shayandeh, Lars Liden, and
  Jianfeng Gao. 2021.
\newblock \href {https://doi.org/10.1162/tacl_a_00399} {Soloist: Building task
  bots at scale with transfer learning and machine teaching}.
\newblock \emph{Transactions of the Association for Computational Linguistics},
  9:807--824.

\bibitem[{Post(2018)}]{sacre-bleu}
Matt Post. 2018.
\newblock \href {https://www.aclweb.org/anthology/W18-6319} {A call for clarity
  in reporting {BLEU} scores}.
\newblock In \emph{Proceedings of the Third Conference on Machine Translation:
  Research Papers}, pages 186--191, Belgium, Brussels. Association for
  Computational Linguistics.

\bibitem[{Qin et~al.(2022)Qin, Welleck, Khashabi, and
  Choi}]{cold_decoding_2022}
Lianhui Qin, Sean Welleck, Daniel Khashabi, and Yejin Choi. 2022.
\newblock \href {https://openreview.net/forum?id=TiZYrQ-mPup} {{COLD} decoding:
  Energy-based constrained text generation with langevin dynamics}.
\newblock In \emph{Advances in Neural Information Processing Systems}.

\bibitem[{Radford et~al.(2019)Radford, Wu, Child, Luan, Amodei, and
  Sutskever}]{radford2019language}
Alec Radford, Jeff Wu, Rewon Child, David Luan, Dario Amodei, and Ilya
  Sutskever. 2019.
\newblock Language models are unsupervised multitask learners.

\bibitem[{Raffel et~al.(2020)Raffel, Shazeer, Roberts, Lee, Narang, Matena,
  Zhou, Li, and Liu}]{raffel2020t5}
Colin Raffel, Noam Shazeer, Adam Roberts, Katherine Lee, Sharan Narang, Michael
  Matena, Yanqi Zhou, Wei Li, and Peter~J. Liu. 2020.
\newblock \href {http://jmlr.org/papers/v21/20-074.html} {Exploring the limits
  of transfer learning with a unified text-to-text transformer}.
\newblock \emph{Journal of Machine Learning Research}, 21(140):1--67.

\bibitem[{Rashkin et~al.(2021)Rashkin, Reitter, Tomar, and
  Das}]{rashkin2021ctrl}
Hannah Rashkin, David Reitter, Gaurav~Singh Tomar, and Dipanjan Das. 2021.
\newblock \href {https://doi.org/10.18653/v1/2021.acl-long.58} {Increasing
  faithfulness in knowledge-grounded dialogue with controllable features}.
\newblock In \emph{Proceedings of the 59th Annual Meeting of the Association
  for Computational Linguistics and the 11th International Joint Conference on
  Natural Language Processing (Volume 1: Long Papers)}, pages 704--718, Online.
  Association for Computational Linguistics.

\bibitem[{Reiser(2004)}]{reiser_scaffolding_2004}
Brian~J. Reiser. 2004.
\newblock \href {https://doi.org/10.1207/s15327809jls1303_2} {Scaffolding
  {Complex} {Learning}: {The} {Mechanisms} of {Structuring} and
  {Problematizing} {Student} {Work}}.
\newblock \emph{Journal of the Learning Sciences}, 13(3):273--304.
\newblock Publisher: Routledge \_eprint:
  https://doi.org/10.1207/s15327809jls1303\_2.

\bibitem[{Roller et~al.(2021)Roller, Dinan, Goyal, Ju, Williamson, Liu, Xu,
  Ott, Smith, Boureau, and Weston}]{roller2020recipes}
Stephen Roller, Emily Dinan, Naman Goyal, Da~Ju, Mary Williamson, Yinhan Liu,
  Jing Xu, Myle Ott, Eric~Michael Smith, Y-Lan Boureau, and Jason Weston. 2021.
\newblock \href {https://aclanthology.org/2021.eacl-main.24} {Recipes for
  building an open-domain chatbot}.
\newblock In \emph{Proceedings of the 16th Conference of the European Chapter
  of the Association for Computational Linguistics: Main Volume}, pages
  300--325, Online. Association for Computational Linguistics.

\bibitem[{Roschelle and Teasley(1995)}]{roschelle1995construction}
Jeremy Roschelle and Stephanie~D Teasley. 1995.
\newblock The construction of shared knowledge in collaborative problem
  solving.
\newblock In \emph{Computer supported collaborative learning}, pages 69--97.
  Springer.

\bibitem[{Ruan et~al.(2019)Ruan, Jiang, Xu, Tham, Qiu, Zhu, Murnane, Brunskill,
  and Landay}]{ruan2019quizbot}
Sherry Ruan, Liwei Jiang, Justin Xu, Bryce~Joe{-}Kun Tham, Zhengneng Qiu,
  Yeshuang Zhu, Elizabeth~L. Murnane, Emma Brunskill, and James~A. Landay.
  2019.
\newblock \href {https://doi.org/10.1145/3290605.3300587} {Quizbot: {A}
  dialogue-based adaptive learning system for factual knowledge}.
\newblock In \emph{Proceedings of the 2019 {CHI} Conference on Human Factors in
  Computing Systems, {CHI} 2019, Glasgow, Scotland, UK, May 04-09, 2019}, page
  357. {ACM}.

\bibitem[{Schick and Sch{\"u}tze(2022)}]{schick2022true}
Timo Schick and Hinrich Sch{\"u}tze. 2022.
\newblock True few-shot learning with prompts—a real-world perspective.
\newblock \emph{Transactions of the Association for Computational Linguistics},
  10:716--731.

\bibitem[{Shuster et~al.(2021)Shuster, Poff, Chen, Kiela, and
  Weston}]{shuster2021retrievalaugmentation}
Kurt Shuster, Spencer Poff, Moya Chen, Douwe Kiela, and Jason Weston. 2021.
\newblock \href {https://doi.org/10.18653/v1/2021.findings-emnlp.320}
  {Retrieval augmentation reduces hallucination in conversation}.
\newblock In \emph{Findings of the Association for Computational Linguistics:
  EMNLP 2021}, pages 3784--3803, Punta Cana, Dominican Republic. Association
  for Computational Linguistics.

\bibitem[{Shuster et~al.(2022)Shuster, Xu, Komeili, Ju, Smith, Roller, Ung,
  Chen, Arora, Lane et~al.}]{shuster2022blenderbot}
Kurt Shuster, Jing Xu, Mojtaba Komeili, Da~Ju, Eric~Michael Smith, Stephen
  Roller, Megan Ung, Moya Chen, Kushal Arora, Joshua Lane, et~al. 2022.
\newblock \href {https://arxiv.org/abs/2208.03188} {Blenderbot 3: a deployed
  conversational agent that continually learns to responsibly engage}.
\newblock \emph{ArXiv preprint}, abs/2208.03188.

\bibitem[{Sinha and Kapur(2021)}]{sinha2021problem}
Tanmay Sinha and Manu Kapur. 2021.
\newblock When problem solving followed by instruction works: Evidence for
  productive failure.
\newblock \emph{Review of Educational Research}, 91(5):761--798.

\bibitem[{Stasaski et~al.(2020)Stasaski, Kao, and
  Hearst}]{stasaski-etal-2020-cima}
Katherine Stasaski, Kimberly Kao, and Marti~A. Hearst. 2020.
\newblock \href {https://www.aclweb.org/anthology/2020.bea-1.5} {{CIMA}: A
  large open access dialogue dataset for tutoring}.
\newblock In \emph{Proceedings of the Fifteenth Workshop on Innovative Use of
  NLP for Building Educational Applications}, pages 52--64, Seattle, WA, USA
  â†’ Online. Association for Computational Linguistics.

\bibitem[{Suresh et~al.(2022{\natexlab{a}})Suresh, Jacobs, Harty, Perkoff,
  Martin, and Sumner}]{suresh2022talkmoves}
Abhijit Suresh, Jennifer Jacobs, Charis Harty, Margaret Perkoff, James~H
  Martin, and Tamara Sumner. 2022{\natexlab{a}}.
\newblock The talkmoves dataset: K-12 mathematics lesson transcripts annotated
  for teacher and student discursive moves.
\newblock \emph{arXiv preprint arXiv:2204.09652}.

\bibitem[{Suresh et~al.(2022{\natexlab{b}})Suresh, Jacobs, Perkoff, Martin, and
  Sumner}]{suresh-etal-2022-fine}
Abhijit Suresh, Jennifer Jacobs, Margaret Perkoff, James~H. Martin, and Tamara
  Sumner. 2022{\natexlab{b}}.
\newblock \href {https://doi.org/10.18653/v1/2022.bea-1.11} {Fine-tuning
  transformers with additional context to classify discursive moves in
  mathematics classrooms}.
\newblock In \emph{Proceedings of the 17th Workshop on Innovative Use of NLP
  for Building Educational Applications (BEA 2022)}, pages 71--81, Seattle,
  Washington. Association for Computational Linguistics.

\bibitem[{Sutskever et~al.(2014)Sutskever, Vinyals, and
  Le}]{sutskever2014sequence}
Ilya Sutskever, Oriol Vinyals, and Quoc~V. Le. 2014.
\newblock \href
  {https://proceedings.neurips.cc/paper/2014/hash/a14ac55a4f27472c5d894ec1c3c743d2-Abstract.html}
  {Sequence to sequence learning with neural networks}.
\newblock In \emph{Advances in Neural Information Processing Systems 27: Annual
  Conference on Neural Information Processing Systems 2014, December 8-13 2014,
  Montreal, Quebec, Canada}, pages 3104--3112.

\bibitem[{Tack and Piech(2022)}]{tack_ai_2022}
Ana{\"i}s Tack and Chris Piech. 2022.
\newblock The {{AI Teacher Test}}: {{Measuring}} the {{Pedagogical Ability}} of
  {{Blender}} and {{GPT-3}} in {{Educational Dialogues}}.
\newblock In \emph{The 15th {{International Conference}} on {{Educational Data
  Mining}}}, page accepted.

\bibitem[{Wen et~al.(2017)Wen, Vandyke, Mrk{\v{s}}i{\'c}, Ga{\v{s}}i{\'c},
  Rojas-Barahona, Su, Ultes, and Young}]{wen2017camrest}
Tsung-Hsien Wen, David Vandyke, Nikola Mrk{\v{s}}i{\'c}, Milica
  Ga{\v{s}}i{\'c}, Lina~M. Rojas-Barahona, Pei-Hao Su, Stefan Ultes, and Steve
  Young. 2017.
\newblock \href {https://aclanthology.org/E17-1042} {A network-based end-to-end
  trainable task-oriented dialogue system}.
\newblock In \emph{Proceedings of the 15th Conference of the {E}uropean Chapter
  of the Association for Computational Linguistics: Volume 1, Long Papers},
  pages 438--449, Valencia, Spain. Association for Computational Linguistics.

\bibitem[{Wollny et~al.(2021)Wollny, Schneider, Di~Mitri, Weidlich, Rittberger,
  and Drachsler}]{wollny2021we}
Sebastian Wollny, Jan Schneider, Daniele Di~Mitri, Joshua Weidlich, Marc
  Rittberger, and Hendrik Drachsler. 2021.
\newblock Are we there yet?-a systematic literature review on chatbots in
  education.
\newblock \emph{Frontiers in artificial intelligence}, 4.

\bibitem[{Xue et~al.(2021)Xue, Constant, Roberts, Kale, Al-Rfou, Siddhant,
  Barua, and Raffel}]{xue2021mt5}
Linting Xue, Noah Constant, Adam Roberts, Mihir Kale, Rami Al-Rfou, Aditya
  Siddhant, Aditya Barua, and Colin Raffel. 2021.
\newblock \href {https://doi.org/10.18653/v1/2021.naacl-main.41} {m{T}5: A
  massively multilingual pre-trained text-to-text transformer}.
\newblock In \emph{Proceedings of the 2021 Conference of the North American
  Chapter of the Association for Computational Linguistics: Human Language
  Technologies}, pages 483--498, Online. Association for Computational
  Linguistics.

\bibitem[{Zhang et~al.(2018)Zhang, Dinan, Urbanek, Szlam, Kiela, and
  Weston}]{zhang2018personalizing}
Saizheng Zhang, Emily Dinan, Jack Urbanek, Arthur Szlam, Douwe Kiela, and Jason
  Weston. 2018.
\newblock \href {https://doi.org/10.18653/v1/P18-1205} {Personalizing dialogue
  agents: {I} have a dog, do you have pets too?}
\newblock In \emph{Proceedings of the 56th Annual Meeting of the Association
  for Computational Linguistics (Volume 1: Long Papers)}, pages 2204--2213,
  Melbourne, Australia. Association for Computational Linguistics.

\bibitem[{Zhang et~al.(2020)Zhang, Sun, Galley, Chen, Brockett, Gao, Gao, Liu,
  and Dolan}]{zhang2020dialogpt}
Yizhe Zhang, Siqi Sun, Michel Galley, Yen-Chun Chen, Chris Brockett, Xiang Gao,
  Jianfeng Gao, Jingjing Liu, and Bill Dolan. 2020.
\newblock \href {https://doi.org/10.18653/v1/2020.acl-demos.30} {{DIALOGPT} :
  Large-scale generative pre-training for conversational response generation}.
\newblock In \emph{Proceedings of the 58th Annual Meeting of the Association
  for Computational Linguistics: System Demonstrations}, pages 270--278,
  Online. Association for Computational Linguistics.

\bibitem[{Zhao et~al.(2016)Zhao, Sinha, Black, and Cassell}]{zhao2016socially}
Ran Zhao, Tanmay Sinha, Alan~W Black, and Justine Cassell. 2016.
\newblock Socially-aware virtual agents: Automatically assessing dyadic rapport
  from temporal patterns of behavior.
\newblock In \emph{International conference on intelligent virtual agents},
  pages 218--233. Springer.

\bibitem[{Zhou et~al.(2018)Zhou, Prabhumoye, and Black}]{cmu_dog_emnlp18}
Kangyan Zhou, Shrimai Prabhumoye, and Alan~W Black. 2018.
\newblock \href {https://doi.org/10.18653/v1/D18-1076} {A dataset for document
  grounded conversations}.
\newblock In \emph{Proceedings of the 2018 Conference on Empirical Methods in
  Natural Language Processing}, pages 708--713, Brussels, Belgium. Association
  for Computational Linguistics.

\end{thebibliography}
\bibliographystyle{acl_natbib}

\clearpage
\appendix

\section{Pedagogical strategy and dialog acts in dialog tutoring}
\label{ap:ped-strategies-da}
\begin{figure}[h!]
    \centering
    \includegraphics[width=\linewidth]{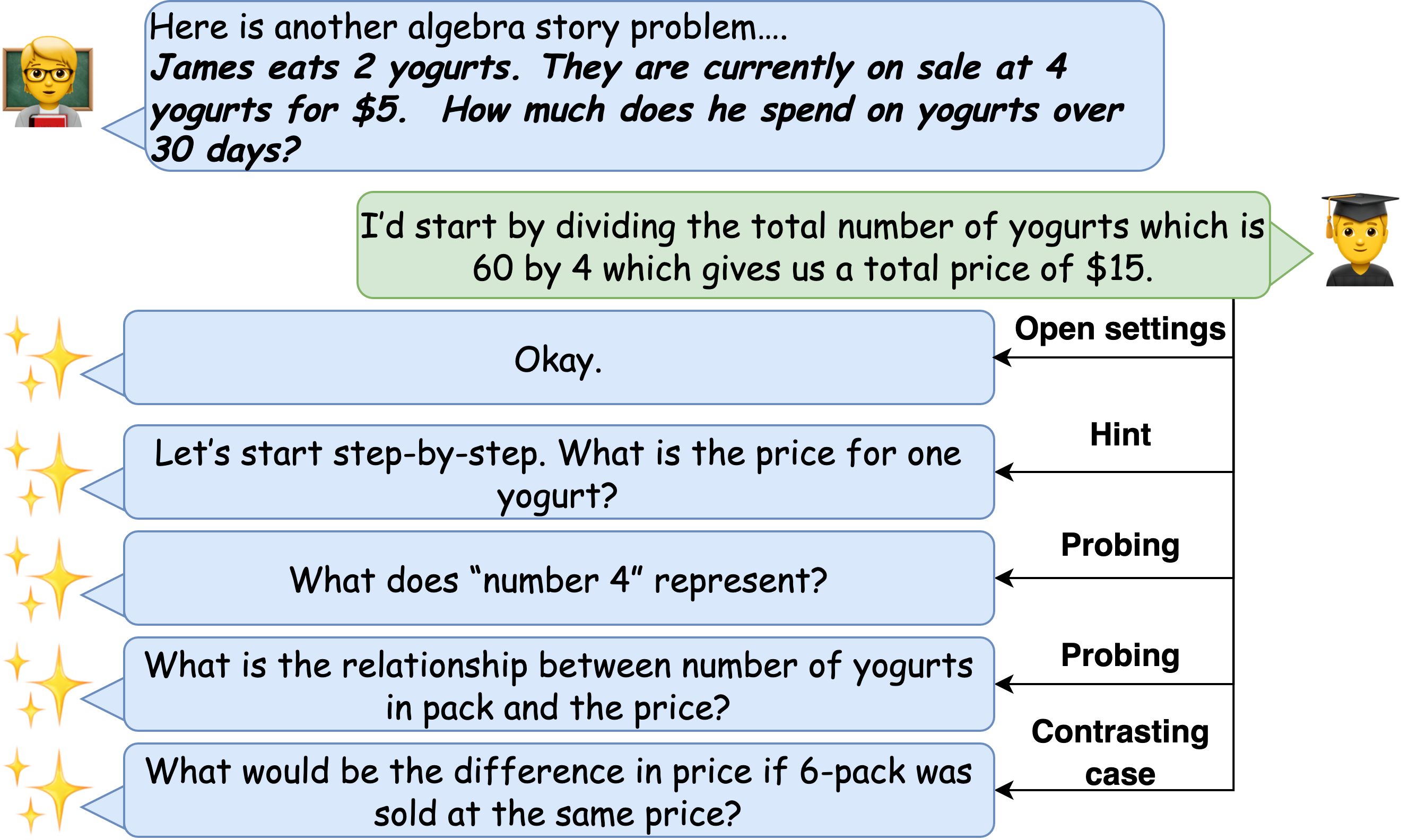}
    \caption{Example dialogue between a tutor and a student solving an algebra story problem. Key questions are: What teacher pedagogical strategies are the best in terms of learning gains of students? How to adapt language models to generate pedagogically valid responses?
}
    \label{fig:dialogue-tutoring-intro1}
\end{figure}
In the context of this paper, we assume that the pedagogical strategy is represented using dialog act annotations. An example of the teacher strategy is providing hints (cf. example in Figure \ref{fig:dialogue-tutoring-intro1}), where a teacher provides helpful support or clarifies goals to the student. Another example is Probing (cf. example in Figure \ref{fig:dialogue-tutoring-intro1}), which prompts students to explain better or reflect on the current solution. CIMA contains five teacher dialog acts - \textit{hint, open-ended question, correction, confirmation, other}. TSCC contains more fine-grained dialog acts such as \textit{eliciting, scaffolding, enquiry, or recap}. 

From a learning science standpoint, pedagogical strategy could be viewed as a global strategy (knowing how to effectively guide students e.g. using questioning or providing contrasting cases) and dialog acts as a specific decision on how this strategy is implemented on the local turn-based level.

\section{Equitable tutoring}\label{sec:equitable}

Although tutoring is typically conceived as a scenario where a subject matter expert works synchronously with one or multiple students and takes interpretive authority, there is increasing empirical evidence supporting the case for incorporating active learning approaches in the classroom \cite{freeman2014active,sinha2021problem}. With collaborative creation of knowledge where teachers position themselves as co-learners and students also take interpretive authority, such approaches are better poised to build classroom equity than monologic educational practices where only one voice (primarily the teacher's) tends to be heard, legitimized and sometimes imposed. Therefore, if we rethink of the goals of education as providing opportunities for students to enter into the workforce with a positive identity about themselves and the subject matter, equitable tutoring via increased student chances to pose ideas, construct knowledge and as a result feel welcomed into the intellectual discussion, holds tremendous promise.

\section{Conversational Goals}\label{appendix-conversational-goals} 
In this work, we studied only 1:1 dialog tutoring settings with a specific focus on the role of a teacher/tutor. We focused on the most commonly used goal of dialog tutoring which is a learning aid to support students' skill development and provide opportunities to learn \cite{wollny2021we}. 

However, teacher-student interactions may entail multiple conversational goals %
that can serve interactional functions (e.g., turn-taking) and interpersonal functions (e.g., rapport-building moves such as self-disclosure, praise, social norm violation). Research in human tutoring and collaborative learning, more generally, has shown that how students manage the task space (consisting of the problem to be solved, e.g., do I agree with the interlocutor's reasoning?) and the relational space (consisting of the interactional challenges and opportunities, e.g., can I disagree without threatening the interlocutor's face?) is critical to learning outcomes \cite{roschelle1995construction,zhao2016socially}. Neural dialog tutoring, which can account for this crucial, but fundamental distinction among conversational goals is challenging.%

\section{Qualitative user evaluation}
\begin{figure}[h!]
    \centering
    \includegraphics[width=\linewidth]{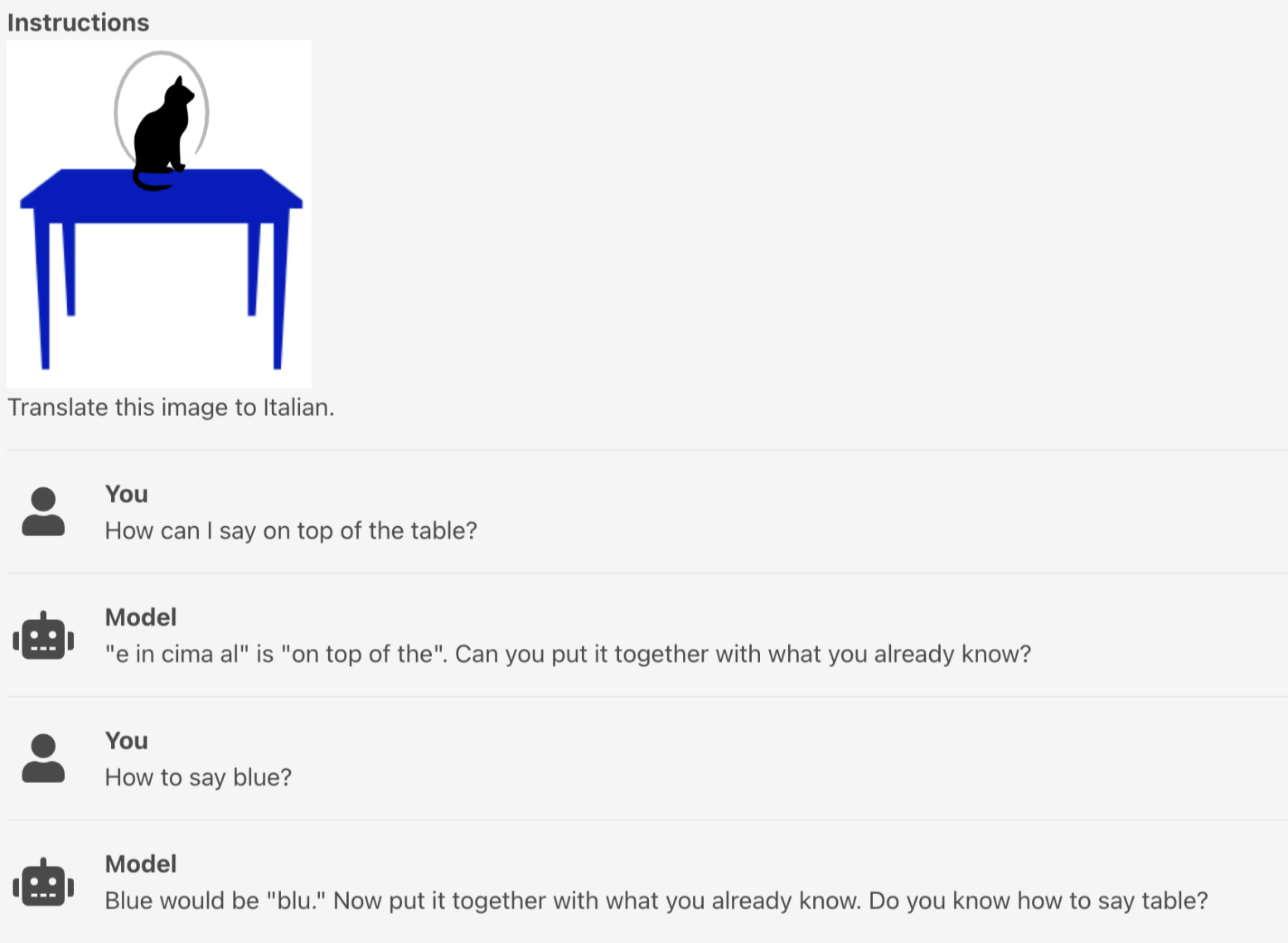}
    \caption{ User interface used for the qualitative evaluation. The interface is based on ParlAI \citep{miller2017parlai} library. 
}
    \label{fig:dialogue-tutoring-interface}
    \vspace{-1.0em}
\end{figure}

The interface used in the qualitative evaluation is shown in Figure \ref{fig:dialogue-tutoring-interface}.

\begin{table*}
\begin{tabularx}{\textwidth}{|X|X|}
\hline
Context & Generated teacher response \\
\hline
T: Please try to fill in the blank in Italian. \newline
S: how do you say next? \newline
T: Remember that  "is next to the" is "e vicino" \newline
S: I don't know how to say purple. Is it viola?
& 
Viola is the word for purple.
\\
\hline
T: Please try to fill in the blank in Italian.  \newline
S: how do you say next? \newline
T: Remember that  "is next to the" is "e vicino" \newline
S: Purple is viola, right?
& 
That is correct. 
\\
\hline
T: Please try to fill in the blank in Italian. \newline
S: how do you say next? \newline
T: Remember that  "is next to the" is "e vicino" \newline
S: I think purple could be violett.
& 
Viola is the word for purple.
\\
\hline
T: Please try to fill in the blank in Italian.  \newline
S: how do you say next? \newline
T: Remember that  "is next to the" is "e vicino" \newline
S: I can't remember the color. Can you give me a hint?
& 
Purple is viola.
\\
\hline
T: Please try to fill in the blank in Italian.  \newline
S: how do you say next? \newline
T: Remember that  "is next to the" is "e vicino" \newline
S: I guess it is e vicino all'albero. Is it correct to start the sentence with the preposition?
& 
The preposition for "purple" is "viola" \newline
\textbf{Understanding - Meaning mismatch}
\\
\hline
T: "Blue" is "blu". Please try to fill in the blank in Italian. \newline
S: But what are the other words? \newline
T: Can you give me your best guess? \newline
S: e di fronte all'albero blu  (\textbf{correct})
& 
Almost, tree is all'albero'. Please try again. \newline
\textbf{Understanding - Incorrect solution assessment}
\\
\hline
T: "Blue" is "blu". Please try to fill in the blank in Italian. \newline
S: But what are the other words? \newline
T: Can you give me your best guess? \newline
S: e di fronte blu all'albero  (\textbf{adjective position swap})
& 
Almost, you need to add blu after "all'albero" to make the word "tree". \newline
\textbf{Understanding - Incorrect translation}
\\
\hline
T: "Blue" is "blu". Please try to fill in the blank in Italian. \newline
S: But what are the other words? \newline
T: Can you give me your best guess? \newline
S: e di fronte al albero blu (\textbf{wrong definite article})
& 
Almost, tree is all'albero not albero. 
\\
\hline
T: "Blue" is "blu". Please try to fill in the blank in Italian. \newline
S: But what are the other words? \newline
T: Can you give me your best guess? \newline
S: e di fronte al cane blu (\textbf{noun swap})
& 
Almost. Al cane is not the correct word for tree.
\\
\hline
T: "Blue" is "blu". Please try to fill in the blank in Italian. \newline
S: But what are the other words? \newline
T: Can you give me your best guess? \newline
S: e dietro all'albero blu" (\textbf{wrong preposition})
& 
The correct answer is e di fronte all'albero blu. \newline
\textbf{Pedagogy - Information reveal}
\\
\hline
\end{tabularx}
\caption{Examples of rephrased conversations from the qualitative study. $T$ refers to a teacher utterance, $S$ refers to a student utterance. Bold text is information for the reader indicating error categories.}
\label{tab:user-study-questions}
\end{table*}

\end{document}